\documentclass[fleqn,10pt]{wlscirep}
\usepackage[utf8]{inputenc}
\usepackage[T1]{fontenc}
\usepackage{float}
\begin{document}
\title{Forecasting trends in food security with real time data}

\author[a]{Joschka Herteux}
\author[b]{Christoph Räth} 
\author[a]{Giulia Martini}
\author[c]{Amine Baha} 
\author[a]{Kyriacos Koupparis}
\author[a]{Ilaria Lauzana}
\author[a,*]{Duccio Piovani}

\affil[a]{World Food Programme, Research, Assessment and Monitoring Division (RAM), Via
Cesare Giulio Viola 68, 00148 Rome, Italy}
\affil[b]{Deutsches Zentrum für Luft-- und Raumfahrt (DLR), Institute for AI Safety and Security, Wilhelm-Runge-Straße 10, 89081 Ulm, Germany}
\affil[c]{World Food Programme Innovation Accelerator, Buttermelcherstrasse 16, 80469 Munich, Germany}
\affil[*]{Corresponding author: Duccio Piovani, duccio.piovani@wfp.org}

\begin{abstract}

Early warning systems are an essential tool for effective humanitarian action. Advance warnings on impending disasters facilitate timely and targeted response which help save lives and livelihoods. In this work we present a quantitative methodology to forecast levels of food consumption for 60 consecutive days, at the sub-national level, in four countries: Mali, Nigeria, Syria, and Yemen. The methodology is built on publicly available data from the World Food Programme’s global hunger monitoring system which collects, processes, and displays daily updates on key food security metrics, conflict, weather events, and other drivers of food insecurity. In this study we assessed the performance of various models including Autoregressive Integrated Moving Average (ARIMA), Extreme Gradient Boosting (XGBoost), Long Short Term Memory (LSTM) Network, Convolutional Neural Network (CNN), and Reservoir Computing (RC), by comparing their Root Mean Squared Error (RMSE) metrics. Our findings highlight Reservoir Computing as a particularly well-suited model in the field of food security given both its notable resistance to over-fitting on limited data samples and its efficient training capabilities. The methodology we introduce establishes the groundwork for a global, data-driven early warning system designed to anticipate and detect food insecurity.
\end{abstract}

\flushbottom
\maketitle

\thispagestyle{empty}

\section*{Introduction}
Conflict \cite{foods11142098, foods11152301}, climate extremes, and soaring food, fertilizer and energy prices \cite{alexander2022high}  on the heels of an incomplete recovery from the COVID-19 pandemic \cite{picchioni2022impact} have created a food crisis of unprecedented proportions \cite{globalfoodcrisis}. The war in Ukraine further complicated the situation with millions of people a step away from starvation. In such an uncertain environment it is incumbent upon humanitarian agencies to deploy effective early warning systems that monitor food security conditions in the most vulnerable countries \cite{globrepfood}.The World Food Programme (WFP) operates in emergency contexts and its operations include direct delivery of food assistance, cash based transfers, nutrition support, and are characterized by rapid response. This is done  by leveraging an extensive logistic network in partnership with other UN agencies, NGOs, government counter parts and local communities, and requires swift assessments of local needs and vulnerabilities. In this context early warning systems serve as the foundation for preparedness and quick response to potential food crises, whether man-made or natural disasters. They help humanitarian organizations better target assistance to where it is needed the most, hence minimizing duplication and waste. Forecasting systems are an integral part of early warning systems, allowing them to anticipate potential hazards and issue alerts in a timely and effective manner.  A robust approach to creating accurate forecasting tools is to leverage modern time series prediction methods that are based on Machine Learning (ML) \cite{lim2021time}. These methods can be applied to a growing number of available data streams, and have proven to be successful across a range of diverse fields. Examples of these fields include monitoring epidemics \cite{polyvianna2019computer,kraemer2019past, tizzoni2012real, scarpino2019predictability, ardabili2020covid, prakash2020analysis}, predicting financial markets \cite{timmermann2018forecasting, elliott2016forecasting,sezer2020financial, sirignano2019universal}, tracking energy consumption \cite{deb2017review} and forecasting weather patterns and climate change impacts \cite{schultz2021can,mudelsee2019trend}.

\begin{figure*}[]
\centering
\includegraphics[width=\linewidth]{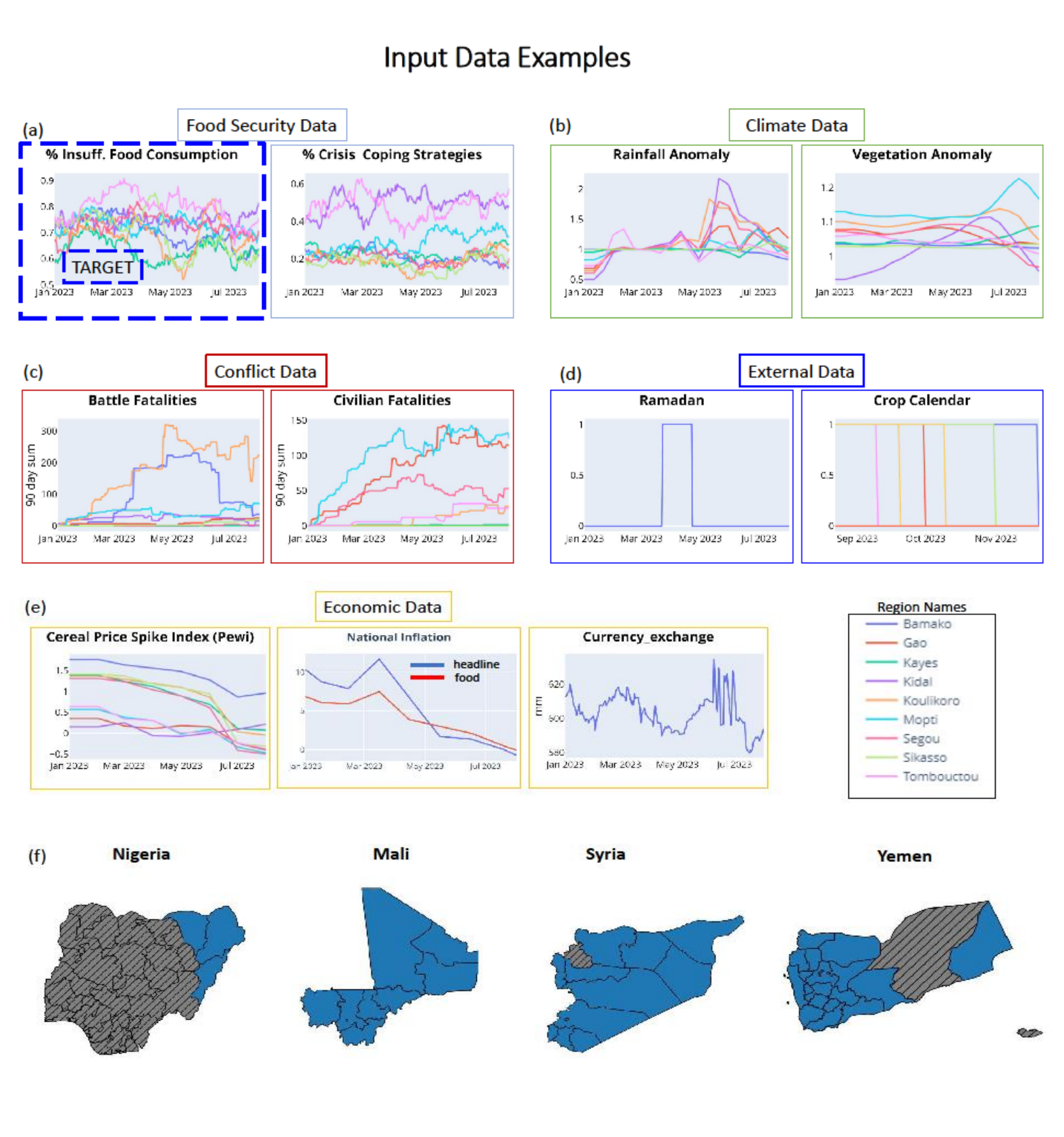}
\caption{\textbf{Input Data}: the figure shows the time series of the data used in constructing the forecasting methodology. The target variable, highlighted by the blue dashed curve in \textbf{(a)}, is the regional prevalence of insufficient food consumption extracted from the Food Consumption Score (FCS). In addition to the historical values of the target, the methodology incorporates predictors coming from another food security indicator, \textbf{(b)} climate , \textbf{(c)} conflict, and  \textbf{(e)} economic data. External datasets with known future values, including crop calendars and Ramadan days, are also considered \textbf{(d)}. While the figures are based on data from Mali, the framework remains consistent across all countries. A detailed breakdown of the data for each country can be found in Table \ref{tab: input data}, with comprehensive information on data sources available in the methods section. \textbf{(f)} The map displays the first administrative level boundaries in the four tested countries, where grey dashed polygons indicate regions that were excluded due to data unavailability.}
\label{fig:FCS}
\end{figure*}

The scientific community has been studying and modeling food security for many decades and recent research has shown that ML can also be used in this context.  By using ML, it is possible to develop complex, data-driven models with minimal feature engineering, thereby avoiding the need for an in-depth understanding of the underlying processes, provided that enough training data is available. This makes ML an attractive candidate for the study of food security, where the problem is complex and reliant on local conditions, making it challenging to generalize any claims that may lead to a knowledge-based model. An ML approach, if well designed, is able to reconstruct the underlying causal pathways without pre-established modelling choices, offering an alternative to the modelling from first principles. There exists a comprehensive body of literature on methods that leverage ML techniques, either standalone or in combination with other approaches.

For instance, the Food and Agriculture Organization of the United Nations (FAO) has investigated long-term forecasts at the country level for several years \cite{world2020state, wanner2014refinements}. In \cite{mwebaze2010causal, okori2011machine} ML algorithms were used to classify households based on their caloric intake in Uganda, from data collected by the Bureau of Statistics. Studies carried out by the World Bank focus on the modeling and forecasting of Integrated Food Security Phase Classification (IPC) phases based on data from the Famine Early Warning Systems Network (FEWS.NET) across fifteen countries \cite{andree2020predicting, wang2022transitions, wang2020stochastic}. The studies include attempts to identify the onset of food crises by building a binary crises indicator from the IPC phases and reconstruct the national share of population living in districts labeled as IPC3+\cite{andree2020predicting}. Moreover in \cite{wang2022transitions} forecasts sub-national IPC phases 1, 2, and 3+ employing a panel vector-auto regression model and using the Least Absolute Shrinkage and Selection Operator (LASSO) technique to pinpoint the most critical drivers of food security. Notably in \cite{krishnamurthy2022anticipating}, based on the same FEWS.NET data, the authors extract significant text-based features from news articles accessed via Factiva. In their work they demonstrate that the text features greatly enhance the predictive capabilities of a Random Forest model, thus exploiting the potential of an abundant data stream that has not been thoroughly explored in the context of food insecurity.

\begin{figure*}%[]
\centering
\includegraphics[width=\linewidth]{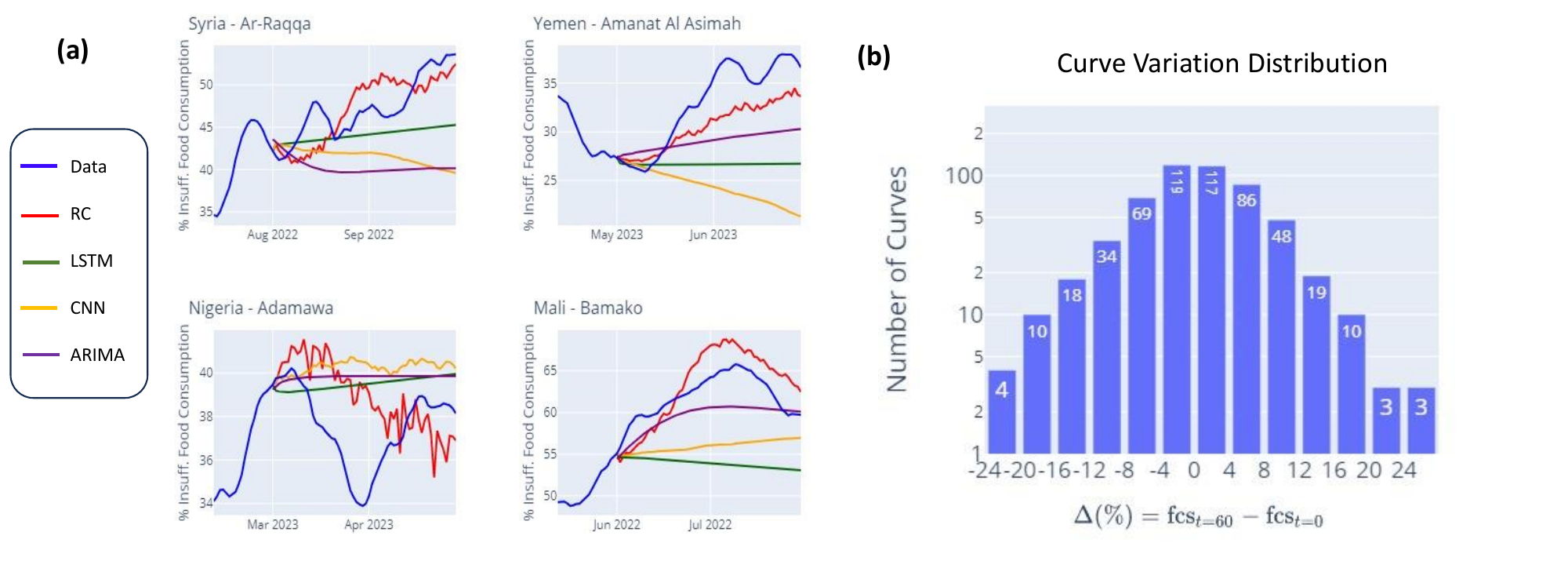}
\caption{\textbf{Forecasts and Data}: 
\textbf{(a)} 60-day forecasts examples generated using the RC, CNN, LSTM, and ARIMA models for four specific sub-national regions: Yemen, Syria, Mali, and Nigeria. In the visual representation, the blue curve represents the actual data, while each of the other curves depicts the prediction of one of the models. \textbf{(b)} The observed distribution of the variation of the prevalence of insufficient food consumption on the 60-day windows used to train and test the algorithms. We can appreciate how the dataset is biased towards curves of small variation.}
\label{fig:pred}
\end{figure*}

Using data from the Living Standards Measurements Study (LSMS) in \cite{lentz2019data}, a LASSO regression was used to forecast food security indicators such as the Food Consumption Score (FCS), the  Household Dietary Diversity Score (HDDS) and the reduced Coping Strategy Index (rCSI), in Malawi. In \cite{westerveld2021forecasting}
an XGBoost algorithm is trained on secondary data in Ethiopia spanning a wide range of available data sets to forecast the future behaviour of the FEWS.NET IPC phases, classified as deteriorations, improvements and no change. A recent study from the French National Centre for Scientific Research (CNRS) has investigated use of Deep Learning approaches to forecast FCS and HDDS indicators as collected annually from the government in Burkina Faso. While researchers from UCLA recently used remotely sensed soil moisture data and food prices to predict changes in food security, using data provided by IPC  in 7 countries \cite{krishnamurthy2022anticipating}. Leveraging real-time data collected by the World Food Programme, in \cite{martini2022machine} authors developed a cross-country prediction model to \textit{Nowcast} FCS and rCSI indicators across sub-national units in multiple countries, while in  \cite{foini2023forecastability} authors trained 30 distinct XGBoost models to construct a 30-day forecasting model for FCS. In this manuscript we used the same target and secondary variables used in \cite{martini2022machine, foini2023forecastability}

The World Food Programme regularly collects data on food security in countries where it operates. The methodology
we present in this manuscript builds upon data sourced from the Real-Time Monitoring (RTM) system, developed by the organization over the last decade and publicly available via HungerMap\textsuperscript{LIVE} \cite{hungermap}. Collected through Computer Assisted Telephone Interviews (CATI)\cite{CATI}, the data provides daily updates on food security indicators at the household level in near 20 countries. One such indicator is the Fpood Consumption Score (FCS), a composite measure that assesses the frequency of consumption of different food groups in the seven days prior to the survey, categorizing households as having poor, borderline, or acceptable food consumption. Households with insufficient food consumption refers to those with poor or borderline food consumption. The aggregation of this variable to the first administrative level in in countries is the target variable for our study. In contrast to other food security monitoring initiatives like IPC, CH, or FEWSNet, WFP's RTM program stands out due to several distinctions. Other programs release country updates at intervals of four or more months while the RTM program provides daily updates throughout the entire year. This unique feature enables real-time monitoring of the dynamic changes in food security levels, setting it apart as a more responsive and timely tool for assessing and addressing evolving challenges. Furthermore, the underlying process to generate  the indicators in the RTM is purely quantitative, and does not rely on domain expert consensus or qualitative observations.

The primary aim of this research is to develop a forecasting methodology that can predict insufficient food consumption, as described in the WFP's RTM system, over a continuous 60-day period. While national development plans can depend on low-frequency data released on fixed dates, the strategic allocation and distribution of resources in emergency scenarios require more frequent updates, such as monthly or weekly, and constant prioritization for optimal impact. Our early warning system is specifically designed for the dynamic and rapidly changing conditions of humanitarian aid operations in crisis settings. The forecasts are updated daily, enhancing their versatility and flexibility, making them continuously useful throughout the year, unlike other methodologies that are restricted by fixed report release schedules. This feature supports frequent re-targeting and optimization, which are crucial for effective humanitarian responses. Given the logistical challenges in hardship conditions where resource distribution can take up to two weeks, our 60-day forecasts are valuable for assessing the risk of deterioration and enabling planning based on updated priorities. Therefore, a 60-day forecasting horizon, with frequent updates, is considered operationally valuable for managing resource distribution programs effectively.

\begin{figure*}%[]
\centering
\includegraphics[width=\linewidth]{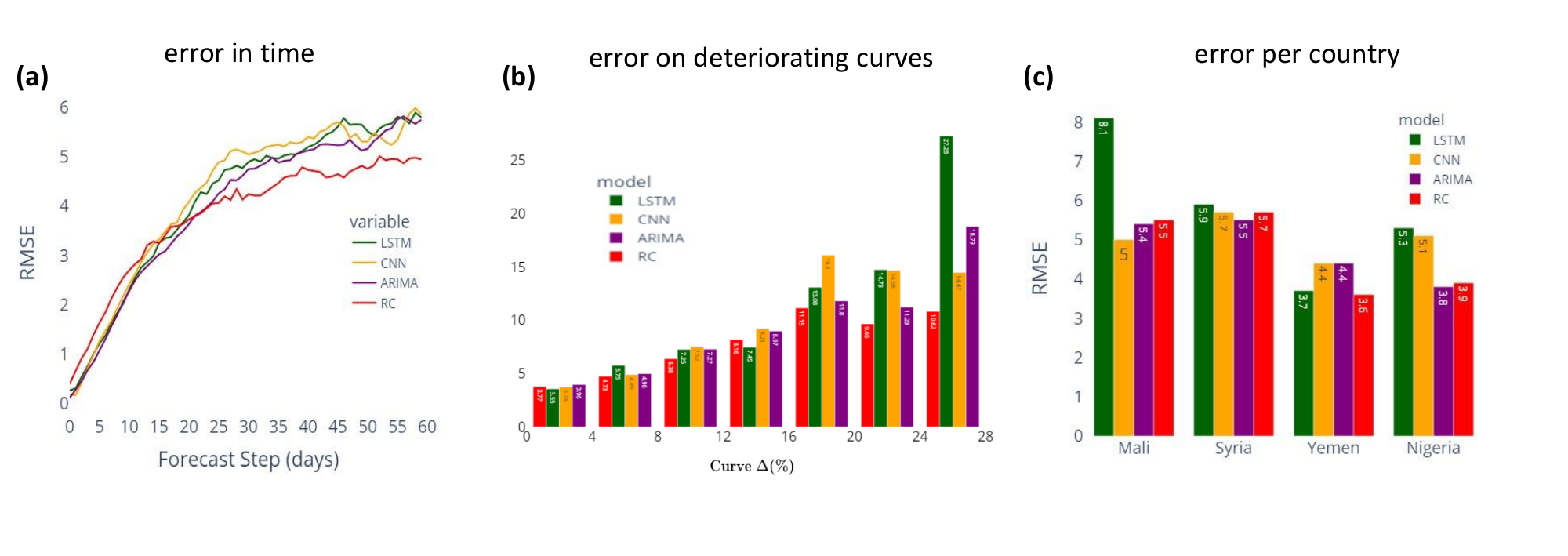}
\caption{\textbf{Forecasting Aggregated Performances}: Performances of the LSTM, CNN and ARIMA models measured in median RMSE, aggregated in three different ways:  Foretasting time step \textbf{a},  variation of the target variable \textbf{b}, per country \textbf{c}. 
Figure \textbf{a} shows how the RC model outperforms the other methodologies after the 15th forested step with an aggregated RMSE at the end of the 60-day window of 4.9 percentage points. In  figure \textbf{b} we see RC tends is consistently the best performer when aggregating on curves according to the variation of their target variable $\Delta = \textbf{fcs}_{60}-\textbf{fcs}_{0}$. This is more evident for high values of $\Delta$, that indicate a sharp increase of levels of insufficient food consumption. The barplot in figure \textbf{c} aggregates the error per country. Despite the fact the RC is among the top performers for every country there is not a clear preferred methodology. This is due to the fact that regional time series belonging to the same country can have very different behaviours and have no clear national characteristic.}
\label{fig:perf1}
\end{figure*} 

We tested out the methodology in four countries, Mali, Nigeria (north east), Syria, and Yemen, at the first sub-national level, using secondary data that represent the key-drivers of food insecurity: conflict, extreme weather events and economic shocks \cite{globrepfood}. The selection of dataset categories for secondary variables in our study draws inspiration from the choices made in previous research \cite{wang2022transitions, westerveld2021forecasting, andree2020predicting, wang2020stochastic, krishnamurthy2022anticipating, lentz2019data, martini2022machine, foini2023forecastability}, with some adjustments based on availability and access convenience. A detailed description is found table \ref{tab: input data}. Our forecasting methodology innovates by leveraging RTM data as the primary target, capitalizing on the unique aspects of the monitoring and incorporating several desirable features that are particularly beneficial for informing humanitarian assistance. In order to increase the transparency and explainability of the model, an aspect that has been identified as crucial in the literature \cite{mcbride2022predicting, baylis2021machine, zhou2022machine}, we further implement a simple method for dynamic feature selection. This allows us to assess the importance of different variables in the model and their impact on the prediction.

The model we identify as the most suitable for forecasting trends in food insecurity is based on Reservoir Computing (RC) \cite{maass02, jaeger2001echo, lukovsevivcius2009reservoir}, a type of recurrent neural network (RNN). Several features make this approach particularly appealing in this context. First, when training the RC, only the parameters of the final layer are "learned," while all other layers are kept constant. This simplification effectively makes RC equivalent to a linear regression, eliminating the need for gradient descent and reducing the number of parameters to learn drastically, speeding up the process and making it suitable for data-constrained contexts. Despite this simplification, the untrained layers and activation functions maintain the model's ability to capture non linear and complex patterns in the data. Additionally, RC allows for the dynamic treatment of secondary variables which are iteratively estimated along with the target variable. The model learns to project these secondary variables into the future, facilitating a straightforward treatment of multidimensional time series. Furthermore, the architecture of RC enables the easy addition of exogenous variables whose future values are known and do not need to be estimated. This feature enhances the model's flexibility and accuracy by incorporating external information directly into the forecasting process. Despite its simplicity RC has proven to be a valid alternative to advanced machine learning prediction techniques particularly in data-scarce contexts as can be found in \cite{chattopadhyay2020, vlachas2019backpropagation,shahi2022prediction} when applied to synthetic data. Specifically our methodology relies on an ensemble of RC models, exploiting the low resource requirements of training RC and stabilizing predictions. This was also recently attempted in\cite{domingo2023anticipating}  where RC is shown to outperform LSTMs on forecasting food prices.

To assess the goodness of our approach, we conducted a comprehensive evaluation by comparing the Root Mean Squared Error (RMSE) of various methodologies, that span deep learning, machine learning and classical statistical model: ARIMA, CNN, LSTM and the XGBoost from a previous attempt \cite{foini2023forecastability}. Inspired by the approach in \cite{westerveld2021forecasting}, we also developed a simple classifier to label future behavior as Deteriorating, No Change, or Improving based on the forecasted values. Our results demonstrate that the RC model outperforms other tested approaches in both types of tasks, and in particular in anticipating dramatic deterioration. This is notable given that curves that show deterioration and severe deterioration are less common in the dataset than the stable curves. For these reasons we believe that the findings presented in this manuscript highlight Reservoir Computing (RC) as a robust option for forecasting applications in the field of food security, where high frequency data is scarce and skewed towards stable behaviour, and for producing early warning signals of impending food crises.

\begin{figure*}[]
\centering
\includegraphics[width=\linewidth]{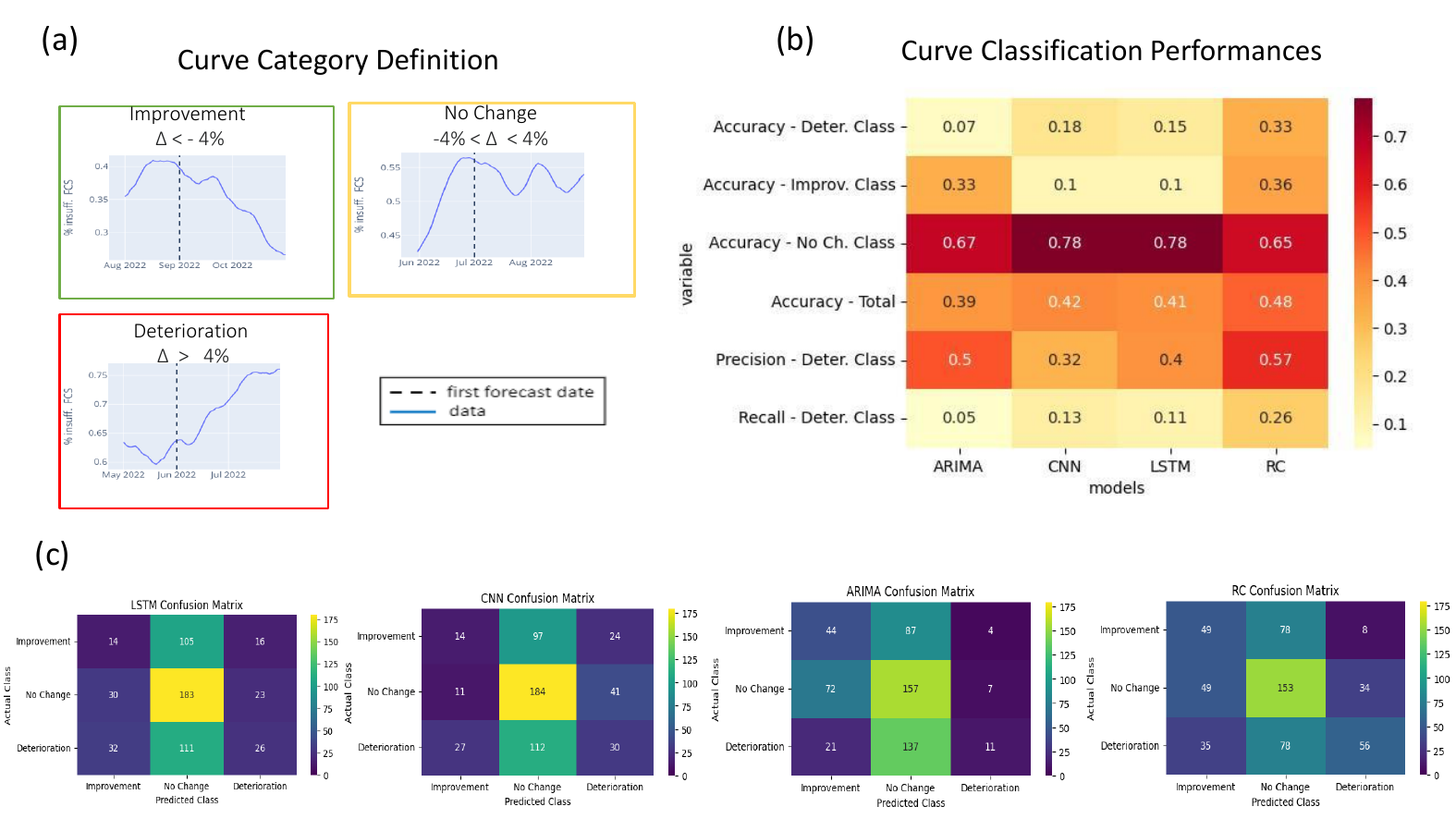}
\caption{\textbf{Classification Performance}: This figures compares the performances of the ARIMA, CNN, LSTM and RC models on their ability to distinguish different classes of behaviour: No Change, Improvement and Deterioration as defined in \textbf{(a)}. The same criterion in applied to actual curves and predicted curves to label the behaviour. \textbf{(b)} Performances metrics on the classification task. Accuracy on singles classes are computed treating the problem as a binary classification \textbf{(c)} Confusion Matrices of all tested methodologies.}
\label{fig:perf2}
\end{figure*}

\section*{Results}

The target variable of our modelling is the time series of the prevalence of households with insufficient food consumption at the first sub-national level. An example is shown in the left panel of figure \ref{fig:FCS}, together with time series examples of a sub sample of the secondary data. Our choice is inspired by the approach used in previous works \cite{foini2023forecastability, martini2022machine} and makes use of available datasets describing conflict, economic shocks, and weather events. More specifically these consist of rainfall and vegetation levels, spikes in market food prices, food and headline inflation, currency exchange rates and number of fatalities due to different kinds of conflict. A detailed description of the input data is provided in the Methods section. Since Islam is the main religion in all four analysis countries, the holy month of Ramadan (fasting) is expected to significantly influence insufficient food consumption (see the Methods section). Therefore, the dates for Ramadan are used as an external input in the RC model, since it is available not only for the past, but also for the prediction period.

To standardize all input variables to the same format of sub-national daily time series we performed some basic pre-processing steps. These include linear interpolation of variables which are only available on monthly basis, aggregating information to the desired sub-national level, averaging intermittent variables over a rolling window as well as de-noising and smoothing all data. Given the stochastic nature of the  RC algorithm, to generate our output we repeat the prediction procedure 100 times with different seeds, with the median of all results then considered the final prediction. 

A comparison between the predicted curves obtained with the different tested methodologies and the actual data is shown in the plots in figure \ref{fig:pred}(a) where the blue curves are the actual data representing the sub-national curves of insufficient food consumption, the colored curves representing the forecasts.In \ref{fig:pred}(b) we can see how the variation of the curves is distributed in the dataset, where $\Delta=\text{fcs}_{60}-\text{fcs}_0$, expressed in percentage points, represents the difference between the last and first values in the forecasting window. A positive $\Delta$ implies that the last value of the window is higher than the first, therefore indicating an increase of food insecurity. On the other hand a negative value indicates an improvement. We refer to curves with positive $\Delta$ as deteriorating curves, while those with a negative $\Delta$ as improving curves. We can see from the figure that the dataset is skewed towards small values of $\Delta$, indicating a majority of stable curves.

\begin{figure}[]
\includegraphics[width=\linewidth]{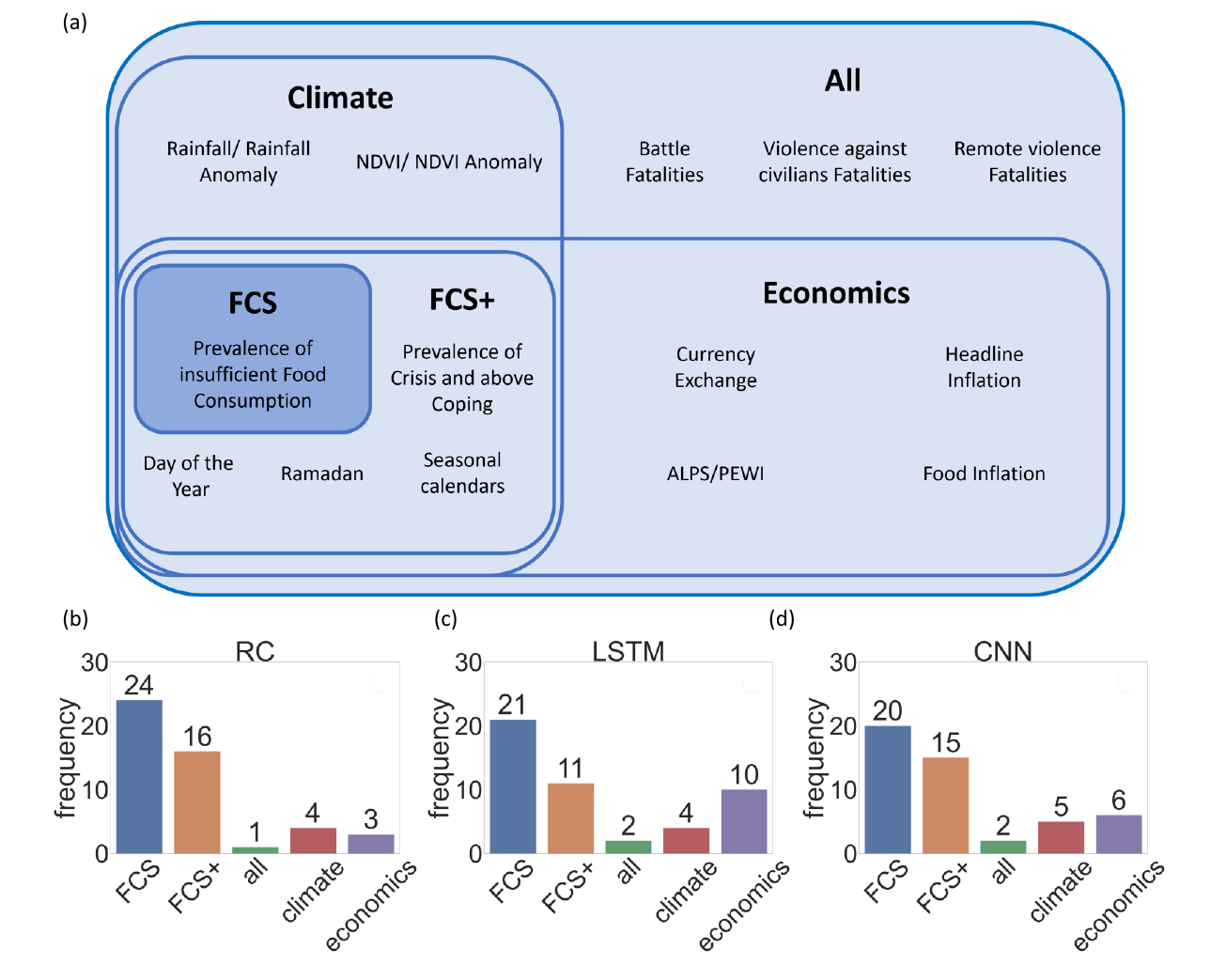}
\caption{\textbf{Feature Selection}: Feature groupings that were considered in the grid search for the RC, CNN and LSTM models to implement a dynamic feature selection.\textbf{(a)} Schema of how the features were classified into 5 different feature groupings. \textbf{(b-d)} frequency with which each feature grouping was selected for in (a) the RC model, (b) the LSTM model and (c) the CNN model.}  
\label{fig:feature groupings}
\end{figure} 

The objective of this work is to test the goodness of the predictions for 60 consecutive days. To ensure a comprehensive comparison among all considered algorithms and to assess their robustness across all seasonal trends, we have evaluated their performances in 12 60-day splits. The forecasts for each split starts at the beginning of every month, covering the period from June 2022 to May 2023. To simulate the use of such methodology in a production environment we employed a walk-forward validation scheme, meaning the model was retrained for every new forecast adding the data of the previous month, therefore expanding the training moving forward in time. At this stage retroactive delays in data updates were not factored in and the models were trained on all available data at present. In a production environment one must take into consideration the update frequency of the secondary variables as well as the target, that may not happen regularly. In section 2 of the supplementary material we further address this issue. For every split the hyper-parameters of the different algorithms were updated and selected through a grid search of the performances on previous splits. Furthermore the algorithms were allowed to choose the subset of input variables that were found to be more informative, i.e. minimized the errors similarly to what was measured in \cite{martini2022machine}.

The aggregate errors are shown in figure \ref{fig:perf1} and \ref{fig:perf2}  and are the result of ca. 70.000 CPU hours of calculations accumulated over several 10.000 runs for each country and model on the German Aerospace Agency's High Performance Cluster. This extensive computation period was essential to thoroughly assess the performances achieved with a vast array of hyper-parameter combinations across all tested splits and countries, ensuring a fair comparison between methodologies. It's noteworthy that the actual training times, on a local standard machine, for an individual hyper-parameter combination typically fall under one minute for all methodologies (See the supplementary material section 5 for more details). As illustrated in figure \ref{fig:perf1}, our analysis of the performances includes a comprehensive study from multiple angles, emphasizing systematic differences in the outputs.

In figure \ref{fig:perf1}\textbf{a} we start by exploring the temporal evolution of the errors, showing the aggregate error for each forecasting step. We can see how the RC model's error rate increases at a slower pace compared to all other considered algorithms. While the RC model is initially outpaced up to the 15th forecasting step, it distinctly becomes the preferable option beyond this point. Figure \ref{fig:perf1}b presents the second type of aggregation for the errors: the degree of change in the data. In the figure we observe that the difference between the performances of the RC and the other tested model widens as the target variable undergoes more substantial change, with RC model outperforming all benchmarks by a clear margin under severely deteriorating conditions. We interpret the general increase across all models in errors for larger data variations as due to the infrequency of such extreme changes, as highlighted in figure\ref{fig:pred} b. Enhancing our approach to better capture these less common types of curves will be the aim of our future work. Figure \ref{fig:perf1}\textbf{c}, shows the results of aggregating the methodologies on different countries, to assess the algorithms' robustness to different sets of available input data sets as well as diverse underlying relationships with the target variable. A complete reference of the available input data sets per country is found in  table \ref{tab: input data}. \ref{fig:perf1}\textbf{c} reveals that both the RC and the ARIMA models emerge as performing better than other methodologies overall, but no clear preferred choice emerges from this aggregation. We believe that this is due to the fact that regional time series of insufficient food consumption do not show any evident national property or characteristics and can assume very different behaviour, even within successive splits. The good performance shown by the simple ARIMA model, frequently  among the top choices in all aggregation, aligns with findings from several other studies \cite{han2023prediction, menculini2021comparing, kobiela2022arima, frausto2021convolutional, domingo2023anticipating}, where the ARIMA model often surpasses or matches the effectiveness of more sophisticated algorithms. As a further evaluation of our proposed approach we compared the performance on 30-day forecasts with those shown in  \cite{foini2023forecastability} on the same dataset. The comparison is found in section 6 of the Supplementary Material where we can see how RC once again stands out the best choice.

\begin{figure}%[]
\centering
\includegraphics[width=0.5\linewidth]{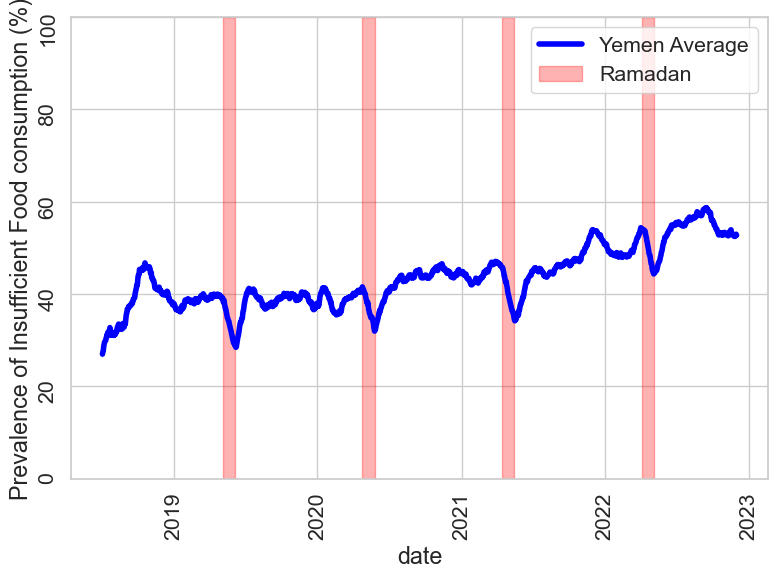}
\caption{\textbf{Ramadan Effect}: Prevalence of people with insufficient food consumption averaged over all available regions of Yemen in blue. Time periods of Ramadan shown by areas shaded red. }
\label{fig:ramadan}
\end{figure}

Inspired by \cite{westerveld2021forecasting}, we classified curves into three categories based on their variation: "No Change" for curves with a variation $|\Delta|<0.04$, "Improvement" for a $\Delta<-0.04$, and "Deterioration" for $\Delta>0.04$. Once again $\Delta = \text{fcs}_{60} - \text{fcs}_0$ is the measure of change seen in the curves. We also applied the same classification criterion to the forecasted curves, turning the regression problem into a classification. From figure \ref{fig:perf2} we can see that the RC model shows superior performance across all metrics with the exception of the accuracy on stable curves. Other methodologies tend to label the majority of curves as being stable therefore falling in the accuracy paradox, where high performances depend on the class unbalance of the dataset rather than on correct prediction. When looking at the minority class \textit{Deterioration}, RC clearly  stands out as the best choice in accuracy, recall and precision. It is worth underlying that all models, including the RC, show a conservative behaviour and tend to forecast a stable outcome. We can see this from the very low levels of the recall on the Deterioration class. This is in line with what was seen in \cite{andree2020predicting} where the False positive rates shown by the methodology are one order of magnitude lower than the False negative rates, and is a limitation of the current methodology that will be addressed in future work.

Finally we have studied what subset of secondary variables produced the lowest error for each split that was tested. In figure \ref{fig:feature groupings} the subsets of input variables are reported  along with the frequency of their selection for every methodology. It is clear that most of the predictive power for all tested methodologies comes from the auto-regressive element, the rCSI indicator and the seasonal external variables Understanding how to optimise the triangulation of the data sources will be the object of future work.  We further analyse this important point in the Training Procedure sub-section of the Methods section.

\section*{Discussion}
This manuscript applies several purely quantitative time series prediction methodologies to the problem of forecasting insufficient food consumption of a horizone of 60 days, based on the data collected by the World Food Programme's real-time monitoring system. In direct comparison with LSTM, CNN and ARIMA we find Reservoir Computing to be the most suitable for this task in terms of performance, resource requirements, robustness and ease of use. We further proved this approach to surpass a previous attempt using XGBoost\cite{foini2023forecastability} (See supplementary material section 6).

The Reservoir Computing algorithm has demonstrated to be a solid framework for the construction of a purely quantitative early warning system, owing to several advantageous features. Its ability to process multidimensional time series and learn complex nonlinear relations in target and features gives it the advantage over simple statistical models like ARIMA. The iterative prediction scheme allows the use of exogenous variables like seasonal and religious calendars where future values are already known as well as make implicit forecasts for those where this is not the case. The relative simplicity of the approach  mitigates the danger of overfitting when compared to the more sophisticated Deep Learning methodologies CNN and LSTM, in line with what was shown in \cite{domingo2023anticipating}. The limited resources needed for the training process allow an ensemble approach, the size of which can be fined tuned and adjusted based on the available computational resources and accuracy requirements of the field and context of application. Furthermore, we saw how the RC methodology was able to accommodate a degree of class imbalance in the training data. Indeed the above benchmark performance shown by the RC model does not depend on the majority of stable curves that form the training set, but is actually more evident in the subset of curves that represent dramatic increases in the levels of insufficient food consumption.  These findings suggest the potential advantages of RC over models of this kind in data-scarce contexts, even when confronted with high-dimensional, noisy, and authentic datasets. The ensemble architecture appears to perform well without requiring specific modifications for large input dimensions, as seen in approaches like Local States \cite{baur21}. Although the comparisons with methodologies based on different target variables is not straightforward, its worth noting that the performances measured with RMSE are in line or better of what presented in other works. In \cite{wang2022transitions} with IPC3+ target the RMSE is of $13\%$, while in \cite{balashankar2023predicting}, the best model performs between $5\%$ and $13\%$ RMSE, although on different time horizons. In \cite{andree2020predicting} the performances vary considerably for different countries up too a maximum of $14.7\%$ RMSE for near term forecasts and to $17.5\%$ on medium term. Other works used different metrics to measure the errors, such as $R^2$, making it more difficult, to compare with what obtained in this work.

The results shown in this manuscript indicate that a forecasting methodology based on the RC algorithm is able to predict 60 days into the future the levels of insufficient food consumption with an average RMSE of 5 percentage points. This increases to around 10 percentage points when only focusing on the subset of curves that show a severe deterioration. From the classification exercise we see that the RC correctly forecasts the type of behaviour once every two forecasts (Total accuracy 0.48), and the number goes down to once every three forecasts when only considering classifying future behaviour as deteriorating of not (accuracy 0.33), or improving or not (accuracy 0.36), according to our definitions. We believe that this is solid starting point for the development of an early warning system based on the RC forecasting algorithm. More work will be done both to improve the general performances and to tailor the methodology to on deteriorating curves.
To achieve this goal we will concentrate our future work on four main pillars: extending the forecasting horizon, tailoring the methodology to detect severe deterioration's of food security levels, developing methods to explain the role of the input variables and implementing the methodology in a production environment. Initial results obtained on 90- day forecasts confirm the RC algorithm as the preferred choice with respect to other benchmarks, despite a general increase of the errors (see Supplementary Material Section 1). Aggregating the underlying data to weekly or monthly time granularity, reducing the number of data points that need to be forecasted and smoothing uninformative fluctuations, seems like a promising possibility. This would also eliminate the need for interpolation and other pre-processing steps on secondary variables, thereby facilitating the deployment and maintanance of the approach. Introducing a custom weighting system in the training process and duplicating data splits which show severe deterioration will be attempted to focus the methodology on deteriorating curves, that are the most important to anticipate. Also testing other and more modern architectures of Reservoir Computing like the Next Generation RC\cite{gauthier2021next} or the Trend-Seasonality decomposition RC \cite{domingo2023anticipating} variation might represent a valid choice to improve overall performances and extend the horizon. In section 2 of the supplementary material we have described a procedure to compute the confidence intervals of the forecasted curves. The confidence intervals may be added considering ±2-standard errors confidence bands. This methodology can be applied to all algorithms including RC. Finally, allowing the algorithms to autonomously select the most informative subset of features for each forecast, as done in this manuscript, marks an initial effort to comprehend the underlying drivers influencing the predictions. A comprehensive analysis of the results and the continued development of this approach will be crucial for enhancing the explanatory power of the methodology.

In conclusion, this research lays the groundwork for a dedicated quantitative tool for modeling and forecasting food insecurity. Our approach complements existing methodologies with its year-round versatility by offering real time updates on future levels of food insecurity. By leveraging the advantageous features of real-time monitoring, the methodology proves particularly valuable for foreseeing and quantifying the impact of events that occur between the releases of other monitoring systems. Our work presents the potential to furnish humanitarian organizations with the capability to rapidly institute early warning systems to preemptively help vulnerable communities. The utilization of extant and expanding datasets in conjunction with the low requirement of computational power renders the proposed approach simple and economical to implement. As such, it could offer much-needed technical support to humanitarian actors in their efforts to anticipate, quantify and promptly address impending food crises.

\section*{Methods}

In this section, we provide an in-depth exploration of the forecasting methodologies employed in our study, along with a comprehensive overview of the input datasets utilized to generate the results discussed in the preceding sections. Particular emphasis is given to the selected Reservoir Computing framework, because of it being less utilised and its central role in this work. In addition, a detailed overview of the training procedures and the hyperparameter selection process for each methodology is shown.

\subsection*{Data}\label{sec:data}
As mentioned in the Introduction and Results sections, the models were trained on data representing the prevalence of food consumption and coping strategies as well as the global key-drivers of food insecurity. The following paragraphs introduce each of the input data sources and the features extracted from them. In table\ref{tab: input data} we can see the availability of the data for each country. 

\medskip
\noindent
\textbf{Prevalence of people with insufficient food consumption}: This measure is built on the Food Consumption Score (FCS), a commonly used food security indicator, and is obtained estimating the percentage of households in a region with an FCS below a country-dependent threshold indicating poor or borderline food consumption. This way a single value between 0 and 100 per region is gained. The Food Consumption Score is collected by WFP both by face to face surveys, which happen approximately once a year and through CATI (Computer Assisted Telephone Interviews) on a daily basis. The score evaluates the amount as well as the nutritional value of a household's food consumption. For some regions not enough data is available and they were thus excluded from the study. This is the target variable of our modelling. 

\medskip
\noindent
\textbf{Prevalence of people using crisis or above crisis food based coping}:
Similar to insufficient food consumption, the prevalence of people using crisis or above crisis food based coping is based on a food security indicator, the reduced Coping Strategy Index (rCSI), which represents a different dimension of the concept. The rCSI is calculated from the results of the same surveys as the FCS and therefore the same methodology is used. Again a prevalence or percentage is defined by counting the number of people in a region scoring below a certain threshold. The rCSI is in particular a measure of the coping strategies adopted by the victims of food insecurity. Examples include limiting the sizes of food portions or the amount of meals per day, borrowing food or relying on help from friends or relatives, and choosing less preferred or less expensive food. Since both indicators stem from the same source, the availability of the data is the same.

\medskip 
\noindent
\textbf{Rainfall and Vegetation}:
Five rainfall and vegetation features were constructed, sourced from WFP's Climate Explorer \cite{climateex} and are originally provided by CHIRPS \cite{chirps}  (rainfall) and MODIS\cite{modis} (vegetation) satellite imagery. These include the amount of rainfall in mm and the NDVI (Normalized Difference Vegetation Index) value, which is a measure for the amount of vegetation in a region. In addition the anomalies of both indicators are included. The anomaly is calculated by dividing the average value of a variable during a period (e.g. one month) by the historical average for the same period in all previous years. To inform the model about the comparison of current and historical seasonal behavior we include the 1- and 3-month anomaly of rainfalland the 1-month anomaly of NDVI. From these WFP derives regional seasonal calendars which mark agriculturally relevant seasons. They can be considered a binary time series of either 0 or 1.  All of these variables are given in 10-day intervals ("dekads") for every region.

\medskip
\noindent
\textbf{Conflict}:
To represent conflict as a driver of food insecurity we use data from the Armed Conflict Location \& Event Data Project (ACLED) \cite{acled} which is a publicly available repository providing real-time and historical data on political violence and
protest events in nearly 100 countries \cite{raleigh2010introducing}. 
We build three features from the fatalities reported for events in an administrative region of the categories Battle, Violence against Civilians and Explosions/Remote Violence by summing each of them in a rolling window over the last 90 days in order to create daily time series for each subnational region.

\medskip
\noindent
\textbf{PEWI Index}: Food prices are an important factor for food security as they affect how much food a household can purchase. The forecasting power of this kind of data has been shown empirically \cite{araujo2012alert}. The ALPS indicator (Alert for Price Spikes) \cite{alps} measures relative spikes in monthly prices by comparing them to estimated values and is calculated monthly as follows
\begin{eqnarray}
    ALPS = ({Price}_t - \hat{Price}_t)/\sigma_\epsilon
\end{eqnarray}
where $t$ gives the month and $\sigma_\epsilon$ describes the historic standard deviation of the residuals (${Price}_t - \hat{Price}_t$ for previous months).
WFP monitors commodity prices in local markets monthly and makes them publicly available through its Economic Explorer \cite{ecoex}. For our model, we only use commodities in the "cereals and tubers" category. The data is averaged over all commodities within this category and across all markets in a given region. This aggregates the data to single time series per first-level administrative unit. 
The ALPS indicator is sometimes also referred to as PEWI (Price Early Warning Indicator).

\medskip
\noindent
\textbf{Macro-Economics}:
To gain a better understanding of the overall situation in a country we also look at several economic indicators, which are available at the country level. We include food and headline inflation to complement our monthly data on food commodity prices. We also use daily effective exchange rates for local currency. All of these indicators can be found in WFP's Economic Explorer \cite{ecoex}. 

\medskip
\noindent
\textbf{Ramadan}:
Ramadan is the month of fasting in the Islamic faith, which is the predominant religion in all four countries considered in this study. The traditional practice of fasting for the whole month can clearly be expected to have a strong influence on food consumption and in  the way it is self-reported. The exact relationship is likely complex, but the impact can empirically be seen in the data. In figure \ref{fig:ramadan} we can see how the target data averaged over sub-national units is affected.  The target time series consistently starts to go down during the time of Ramadan and recovers to roughly the same level afterwards in a similar time frame. These dips are among the most noticeable features of the time series. To accommodate this, we include the occurrence of Ramadan as a binary variable. If a date lies during Ramadan, this feature has a value of 1, otherwise it is 0. Since we know even for future dates if they occur in Ramadan, this feature is processed separately as external data. It is available during prediction and does not need to be inferred. 

\medskip
\noindent
\textbf{Day of the year}:
To give the models the ability to learn seasonal effects as well as possible we include a feature to represent the day of the year as a number ranging from 1 to 365. Similarly to Ramadan this feature is of course known for the future and can be used as an external variable.

\subsection*{Algorithms}
For this study we have investigated a number of forecasting methodologies ranging from the simple statistical model ARIMA (Autoregressive Integrated Moving Average) to the Deep Learning techniques CNN (Convolutional Neural Network) and LSTM (Long Short-Term Memory) where Reservoir Computing (RC) can be understood as a hybrid between the two extremes. 

\medskip
\noindent
\textbf{Ensemble Reservoir Computing Model}
Reservoir Computing is a simplified Recurrent Neural Network(RNN) algorithm known to be easy to train and to be well-suited for problems with limited amounts of data, where it outperforms other state-of-the-art prediction algorithms\cite{chattopadhyay2020,vlachas2019backpropagation,shahi2022prediction}. It is able to process multidimensional inputs as well as past values through its untrained hidden layer called the reservoir and use them to iteratively predict multidimensional time series. As an RNN it is well suited for temporal, sequential data. Because only the last readout layer is explicitly trained through a simple linear regression, it is still considered less complex than typical Deep Learning methodologies like LSTM. To utilize the low resource requirements of RC and stabilize any random fluctuations caused by the initialization of the untrained network weights we implemented a simple ensemble scheme for the model by training 100 separate models with the same hyperparameters similar to what has been applied in \cite{domingo2023anticipating}. For each single model only the seed for the random creation of the reservoir network is changed. The final output of the ensemble model is derived by calculating the median of the compilation of single outputs.

The variant of RC we implement for each single model is based on the original version of RC known as Echo State Network (ESN) \cite{jaeger2001echo}. An ESN is a Recurrent Artificial Neural Network consisting of three layers: an input layer, a hidden reservoir layer and a read-out. While the input layer is simply responsible for distributing the incoming information in the network and the function of the readout is to shape the final output, the reservoir's purpose is firstly the projection of the data to a high-dimensional, nonlinear space and secondly to serve as the memory of the model.
The dynamics of the ESN center around the reservoir state $\textbf{r}_t \in \mathbb{R}^{N}$, which develops according to the update equation 

\begin{eqnarray} 
\textbf{r}_{t+1} = \tanh(\textbf{A}\textbf{r}_{t} +  \textbf{W}_{in} \textbf{x}_{t}) 
\label{eq:simple updating} 
\end{eqnarray}  
where the adjacency matrix $\textbf{A} \in \mathbb{R}^{N \times N} $ represents the network, $\textbf{x}_{t} \in \mathbb{R}^{d_{x}}$ is the input fed into the reservoir and $\textbf{W}_{in} \in \mathbb{R}^{d_{x} \times N}$ is the input matrix. The hyperbolic tangent is a standard choice for the activation function. Since the state of the reservoir is a function of all the past inputs it received, it retains some information about the past if they are provided in sequential order. This is what makes ESN and RC in general well suited for time series analysis and prediction. We create $\textbf{A}$ as a sparse random network since ow connectivity has been found to be advantageous\cite{griffith2019forecasting}. The weights of the network are then drawn uniformly from $[-1,1]$ and afterwards rescaled to fix the spectral radius $\rho$ to some fixed value. $\rho$ is a free hyper-parameter. We chose $\textbf{W}_{in}$ to be also sparse, in the sense that every row has only one nonzero element. This means every reservoir node is only connected to one degree of freedom of the input \cite{lu2018attractor}. The nonzero elements are drawn uniformly from the interval $[-1,1]$ and then rescaled with a factor $s_{input}$, which is another free hyperparameter called input strength. 

From this we can then compute the output $\textbf{y}_{t} \in \mathbb{R}^{d_{y}}$. For our predictive model we train the ESN to approximate $\textbf{y}_{t} \approx \tilde{\textbf{x}}_{t+1}$. Where $\tilde{\textbf{x}}$ is not just the target data, but also all of the secondary data that is not known for the future. This way the model also learns to estimate the secondary variables and project them into the future. The readout is characterized by  
\begin{eqnarray} 
\textbf{y}_{t} = \textbf{W}_{out}\tilde{\textbf{r}}_{t} 
\label{eq:general readout} 
\end{eqnarray}  
where $\tilde{\textbf{r}} = \{r_1,r_2,...,r_N,r_1^2,r_2^2,...,r_N^2\}$. This is a commonly used \cite{lu2017reservoir} nonlinear transformation of $\textbf{r}_{t}$  which additionally serves to break the antisymmetry the equations would otherwise have \cite{herteux2020breaking}. The readout matrix $\textbf{W}_{out} \in \mathbb{R}^{d \times \tilde{N}}$ is the only part of the ESN that is trained. This is done via simple Ridge Regression \cite{hoerl1970ridge} controlled by the regularization parameter $\beta$. 
The resulting model is as a first step capable of forecasting the target variable as well as the required secondary variables one day into the future. To do so the time series of past input data is fed into the reservoir sequentially. During this process the reservoir stores information about past and present values of these variables. The readout can then be used to map from the reservoir to the prediction.
For the purpose of predicting time series of any length the model is used as a "closed loop". The output is fed back into the model allowing the sequence of input data to continue. This way the forecast can be extended to any length, although the error tends to grow with every iteration, thus effectively limiting the reasonable time window. Secondary variables have to be projected into the future in the same way for the procedure to work. Since this creates an additional source of error, we developed a way to include future values of the data in case they are known. For some features like Ramadan or growth seasonality, we do not train the model to forecast the values. Instead, the known future values are used as input during the respective prediction steps. An avenue for future research and an improved feature selection could be the inclusion of independent forecasts for secondary variables like rainfall and conflict, since the RC model is not optimized to predict their behavior.

\medskip
\noindent
\textbf{ARIMA}:
The Autoregressive Integrated Moving Average (ARIMA) is a simple statistical model  designed to predict future behavior of a time series solely based on its past values, without incorporating additional variables. In this model, each time step is presumed to have a linear dependence on the preceding 
$p$ values of the time series, along with 
$q$ error terms representing white noise. To address non-stationary time series, the data undergoes differencing 
$d$ times. While these assumptions limit the model's ability to capture complex nonlinear relationships, ARIMA is recognized for its robustness, efficiency, and ease of training. Consequently, it remains widely applied in various domains such as financial time series \cite{kobiela2022arima} and epidemiology \cite{benvenuto2020application}. 

\medskip
\noindent
\textbf{Convolutional Neural Netorks (CNN)}: a widely recognized category of Artificial Neural Network (ANN) particularly renowned for image recognition tasks. 1-dimensional CNNs have been successfully applied in time series prediction tasks even though they are not primarily designed for sequential data. These networks can incorporate multiple hidden layers and are able to learn temporal dependencies between the target data, its past values and secondary data, which they achieve via an advanced backpropagation-based training scheme. Our model is based on the keras implementation of CNN and consists of a variable number of 1D Convolution layers with ReLu activation functions each followed by a 1D Max Pooling layer. The number of filters, the kernel size and the pool size are among the free hyper-parameters of the architecture which are selected through the grid search.
The readout is comprised of two Dense layers, the first employing ReLu activation functions and a variable number of units followed by a linear layer that maps to the output.
In contrast to the iterative scheme utilized by RC, the model takes a variable number of past values of the employed features as input, and directly outputs a vector of predicted values. During training, the model parameters are trained by minimizing the Mean Squared Error (MSE) using the Adam optimizer for 200 epochs with a variable learning rate. Early Stopping was employed, using 20\% of the training data for validation, to halt the training once the validation error fails to improve for five consecutive epochs.

\medskip
\noindent
\textbf{Long Short-Term Memory (LSTM)}: known as the most widely adopted variant of Recurrent Neural Network, it is part of the larger category of ANNs where it stands out by its sophisticated mechanism to deal with long-range temporal dependencies through selectively retaining past information with the help of its forget gate \cite{hochreiter1997long}. The parameters of this model are trained via backpropagation through time (BPTT). While this is the most sophisticated algorithm we tested in this study, it is also the hardest to train and to fine-tune and carries the highest risk of overfitting. Similarly to our approach with CNNs we built our LSTM model using keras. The architecture consists of two LSTM layers connected by a Repeat Vector Layer. The activation functions used are of the type ReLU while the number of units in both layers is a free hyperparameter subject to tuning in the grid search. The same is true for the dropout rate which we apply to both layers on the input as well as the recurrent connections. The readout of the model consists of a Time Distributed Dense layer mapping to a vector of predicted values. We only apply the model to differenced data as opposed to making this a part of the hyperparameter tuning as in the RC and CNN model due to the fact that we saw a complete inability to learn the original data without differencing with the LSTM in preliminary trials. The training of the LSTM model mirrors that of the CNN model using an Adam optimizer and Early Stopping. 
\begin{figure}%[ht]
\centering
\includegraphics[width=\linewidth]{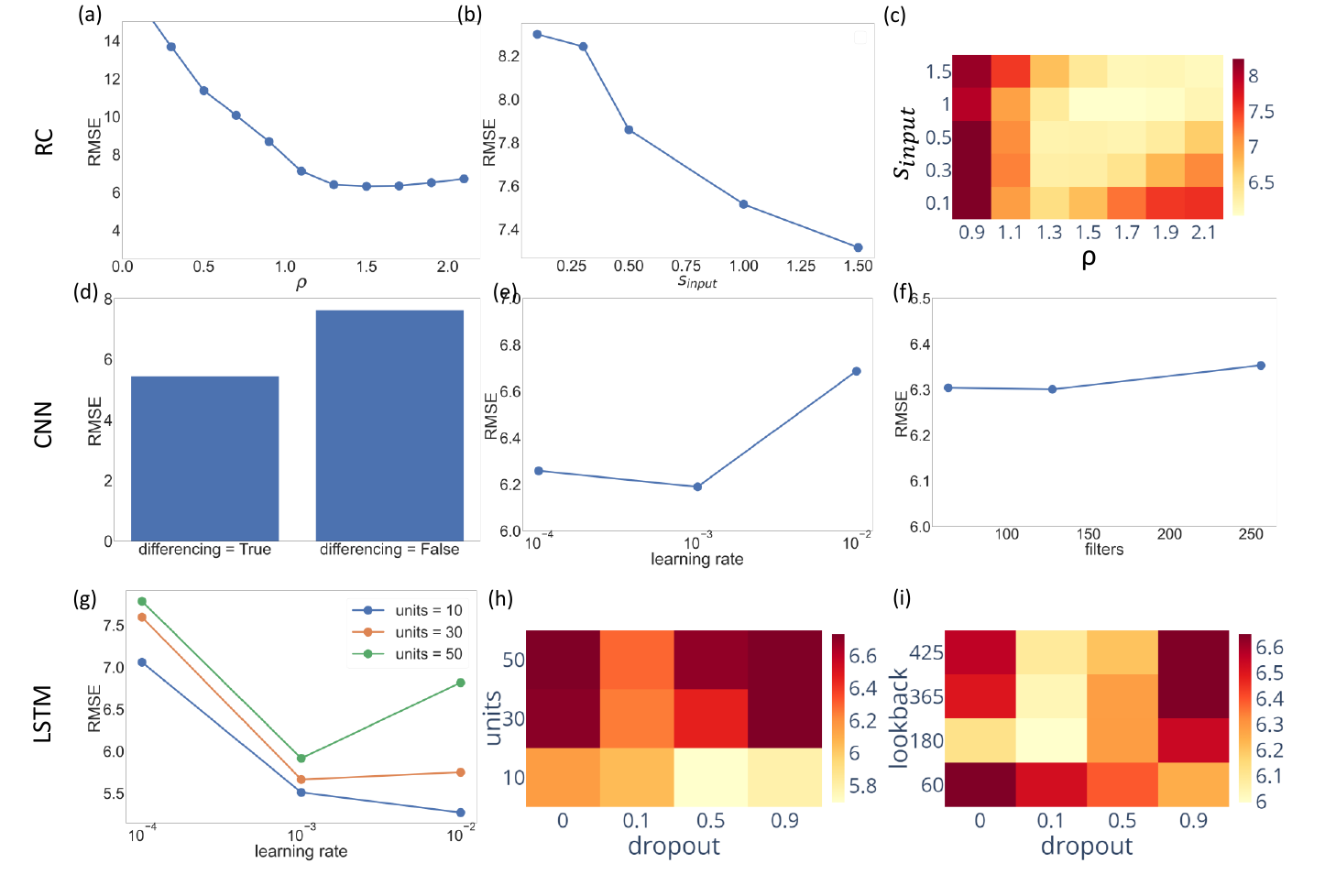}
\caption{\textbf{Grid Search Results}: Additional results on the influence of different hyperparameters derived from the grid search on Yemen. The value shown is always the median RMSE over all splits and unspecified hyperparameters. \textbf{(a-b):} RC results for spectral radius and input strength, which are often considered the most important hyperparameters in RC. Results are separated into runs with and without differencing to showcase varying behavior. \textbf{(d-f):} CNN results on differencing, learning rate and number of filters. \textbf{(g-i):} LSTM results on learning rate, number of units, dropout rate and length of lookback.}
\label{fig:grid search results}
\end{figure}
\subsection*{Training Procedure}
To ensure a fair comparison of all models and determine their optimal configurations, we conducted an extensive grid search on a high-performance cluster for each model and country. In order to simulate real-life scenarios, we implemented a walk-forward validation scheme across the 12 splits under consideration covering a full year.
As a preprocessing step all input data was resampled to a daily resolution and to eliminate isolated missing values linear interpolation was used. A trailing 10-day moving average was applied to smooth the data and reduce the amount of noise. For every prediction, we record the RMSE and the utilized hyperparameters. The result is a rich dataset that enables an in-depth analysis of our models' behavior and the significance of various hyperparameters.

Results directly derived from the grid search are shown in figure \ref{fig:grid search results}. The importance of the spectral radius and the input strength in RC is confirmed in agreement with the literature. For CNN we can see that differencing improves the performance, while other hyperparameters like e.g. the number of filters only have a small impact. For LSTM we identify the learning rate, the number of units, the dropout rate and the length of lookback as having a strong impact. We can see that it is generally impossible to define a single hyperparameter as good or bad since they are usually interdependent. Nevertheless, the fact that high dropout rates are favored in some cases in LSTM indicates a tendency towards overfitting.  Overall, it is observed that  RC stands out as the model for which it is easiest to identify generally reasonable hyperparameter configurations that yield satisfactory performance across all splits. This implies that satisfactory forecasts could even be made with less frequent retuning of the model's hyperparameters.

Within this framework we implement a walk-forward optimization procedure. This technique is commonly used in time series analysis and forecasting to assess the performance of a predictive model. The idea is to train and test the model sequentially over time, moving forward in a step-by-step manner. This process allows the model to be evaluated on new data that becomes available as time progresses.
The process can be separated into the following steps, which are carried out for each model on each country:
\begin{enumerate}
\item \textbf{Selecting a Split}:
A point in the the time series is chosen as the cutoff between training and evaluation. The data before this point is considered to be the training data. The data after this point is considered test data.
\item \textbf{Training the Model}:
The model is trained on the training data.
\item \textbf{Evaluating the Model}:
A forecast is created with the trained model on the time period after the cutoff and compared to the test data where the RMSE is calculated between forecast and test data.
\item \textbf{Walk Forward}: A new split with a new cutoff point further ahead in time is selected and the process is repeated.
\end{enumerate}
This strategy can be used to evaluate the model with the best hyperparameters in the following way: the process is carried out for each combination of hyperparameters considered. To evaluate the performance and simulate an real-life application with regular retraining of the model hyperparameters for each split are chosen based on their aggregate performance across all preceding splits, where the prediction period ended before the beginning of the current one. This prevents any mixing of training and test set. The metric used to evaluate this aggregate performance is the median RMSE over these splits. The choice of the median RMSE over the mean RMSE was motivated by the fact that we found usage of the latter to be more likely to lead to mean-reverting, flat predictions. Conversely, using only a smaller amount of previous splits to select the hyperparameters proved to be too unstable given the heterogenous and noisy nature of the data where the best performance in a single split can often be an outlier. 
With the optimal knowable configuration for a given split we can properly evaluate by looking at the performance of the respective version of the model on the respective time period. Using the differences or returns of the prevalence of insufficient food consumption as target instead of the raw values proved to be particularly important for the LSTM and CNN models which were otherwise mostly unable to successfully forecast. For this reason we added this as an option for RC as well while for ARIMA differencing is a already a part of the regular setup.
 
Beyond the configuration of the algorithms themselves we also investigated a noteworthy modification of the input data. Because the importance of each feature as well as their information content and noise-levels can vary drastically from country to country and even between splits, we implemented a method to select the most useful features for each prediction in a way that could be used in application. In order to achieve this we introduced a hyperparameter for different groups of features in the grid search which makes it possible to optimize it together with the model configuration. The five groups we define are called "FCS", "FCS+", "climate", "economics" and all ranging from including only the target data in "FCS" to the full set of features in "all". Exactly which features are part of the different groups is visualized in figure \ref{fig:feature groupings}a. In figure \ref{fig:feature groupings}b-d we present the frequency with which each of the groupings is selected by the three algorithms RC, LSTM and CNN. The smaller groups of "FCS" and "FCS+" are selected with the highest frequency indicating that the autoregressive component is the most generally predictive one. However, the larger groups especially "climate" and "economics" do make their appearance showing that they are indeed informative at specific instances.
This method was not applied for ARIMA, which does not use secondary data.\\
For all models a large preliminary grid search was carried out on the example of Yemen to identify the most relevant hyperparameters. Based on the results the ranges and fixed parameters were selected as shown in tables \ref{tab: RC HPO},\ref{tab: LSTM HPO},\ref{tab: CNN HPO} and \ref{tab:ARIMA HPO}. While the hyperparameters and their ranges generally differ for each model, our selection focused on those with the highest impact on performances while keeping the overall resource requirements of the grid search similar for RC, CNN and LSTM.

\section*{Data Availability}
The data used in this study is available in public repository https://github.com/ducciopiovani/FamPredAI

\section*{Code Availability}
The code used in this study is available in public repository https://github.com/ducciopiovani/FamPredAI

\bibliography{literature}

\section*{Acknowledgements}

We would like to thank Arif Husain for the continuous support and guidance. We are also grateful to Elisa Omodei and Pietro Foini for sharing their data and their knowledge on the XGBoost approach. We also want to express our gratitude to Will Fein from Microsoft AI for Good for his suggestions that improved our training procedure. This work was supported by the World Food Programme Innovation Accelerator.

\section*{Author contributions}

C.R. conceived and supervised the research, D.P. and J.H. designed the study. J.H., I.L. and D.P.  conducted the calculations. J.H., C.R, G.M, A.B, K.K, I.L and D.P. interpreted and evaluated the findings. J.H., C.R, G.M, A.B, K.K, I.L and D.P. wrote and reviewed the manuscript. 

\section*{Competing interests}
All authors declare no competing interests.

\section*{Correspondence}

Correspondence and requests for materials should be addressed to Duccio Piovani

\begin{table}%[tbhp]
\centering
    \centering
    \begin{tabular}{| l | l | l | l | l | l | l |}
        \hline
        Variable & Yemen & Syria & Mali & Nigeria & frequency & spatial resolution\\
        \hline
        insufficient food consumption  & \checkmark & \checkmark & \checkmark & \checkmark & daily & sub-national \\
        \hline
        food based coping  & \checkmark & \checkmark & \checkmark & \checkmark & daily & sub-national \\
        \hline
        rainfall  & \checkmark & \checkmark & \checkmark & \checkmark & dekad & sub-national \\
        \hline
        rainfall 1 month anomaly & \checkmark & \checkmark & \checkmark & \checkmark  & dekad & sub-national  \\
        \hline
        rainfall 3 months anomaly & \checkmark & \checkmark & \checkmark & \checkmark  & dekad & sub-national  \\
        \hline
        NDVI  & \checkmark & \checkmark & \checkmark & \checkmark & dekad & sub-national \\
        \hline
        NDVI 1 month anomaly & \checkmark & \checkmark & \checkmark & \checkmark  & dekad & sub-national  \\
        \hline
        ALPS/PEWI & \checkmark & \checkmark & \checkmark & \checkmark  & monthly & sub-national  \\
        \hline
        food inflation &   &  & \checkmark & \checkmark  & monthly & national  \\
        \hline
        headline inflation &   &  & \checkmark & \checkmark  & monthly & national  \\
        \hline
        currency exchange official & \checkmark & \checkmark & \checkmark & \checkmark  & daily & national  \\
        \hline
        currency exchange unofficial & \checkmark &  &   &   & daily & national  \\
        \hline
        battle fatalities & \checkmark & \checkmark & \checkmark & \checkmark  & daily & sub-national  \\
        \hline
        violence against civilians fatalities & \checkmark & \checkmark & \checkmark & \checkmark  & daily & sub-national  \\
        \hline
        remote violence/explosions fatalities & \checkmark & \checkmark & \checkmark & \checkmark  & daily & sub-national
        \\
        \hline
        Seasonal calendars & \checkmark & \checkmark & \checkmark & \checkmark  & dekad & sub-national
        \\
        \hline
        Ramadan & \checkmark & \checkmark & \checkmark & \checkmark  & daily & national
        \\
        \hline
        day of the year & \checkmark & \checkmark & \checkmark & \checkmark  & daily & national
        \\
        \hline
        Number of input variables & 266 & 160 & 114 & 360 &  & 
        \\
        \hline
    \end{tabular}
    \caption{\textbf{Input Data}: Variables used in the model for each of the four pilot countries. Time resolution of the data is indicated under frequency. The level of spatial aggregation is shown under spatial resolution. The last row shows the total number of input variables used in each model, which is summed over all sub-national time series available for the specified features. }
    \label{tab: input data}
%\end{*}
\end{table}

\begin{table}
    \centering
    \begin{tabular}{| l | l | l |}
        \hline
        Parameter &  range \\
        \hline
        $\rho$ & 0.3, 0.5, 0.7, 0.9, 1.1, 1.3, 1.5, 1.7, 1.9, 2.1 \\
        \hline
        $\beta$ & $10^{-5}$, $10^{-3}$, $10^{-1}$,10, 100 \\
        \hline
        $s_{input}$ & 0.1, 0.3, 0.5, 1.0, 1.5 \\
        \hline
        features & FCS, FCS+, economics, climate, all \\
        \hline
        differencing & True, False\\
        \hline
    \end{tabular}
    \caption{\textbf{RC Grid Search Parameters}: ranges of RC hyperparameters to be searched during tuning procedure.}
    \label{tab: RC HPO}
\end{table}

\begin{table}
    \centering
    \begin{tabular}{| l | l | l |}
        \hline
        Parameter &  range \\
        \hline
        learning rate & 0.0001, 0.001, 0.01 \\
        \hline
        lookback & 60, 180, 365, 425 \\
        \hline
        units & 10, 30, 50\\
        \hline
        dropout rate & 0., 0.1, 0.5, 0.9 \\
        \hline
        features & FCS, FCS+, economics, climate, all \\
        \hline
    \end{tabular}
    \caption{\textbf{LSTM Grid Search Parameters}: ranges of LSTM hyperparameters to be searched during tuning procedure.}
    \label{tab: LSTM HPO}
\end{table}

\begin{table}
    \centering
    \begin{tabular}{| l | l | l |}
        \hline
        Parameter &  range \\
        \hline
        learning rate & 0.0001,0.001,0.01 \\
        \hline
        lookback & 60, 180, 365, 425 \\
        \hline
        kernel size & 2, 5, 10, 15 \\
        \hline
        kernel size & 64, 128, 256\\
        \hline
        pool size & 2, 4, 6\\
        \hline
        layers & 1, 2 \\
        \hline
        features & FCS, FCS+, economics, climate, all \\
        \hline
        differencing & True, False\\
        \hline
    \end{tabular}
    \caption{\textbf{CNN Grid Search Parameters}: ranges of CNN hyperparameters to be searched during tuning procedure.}
    \label{tab: CNN HPO}
\end{table}

\begin{table}
    \centering
    \begin{tabular}{| l | l | l |}
        \hline
        Parameter &  range \\
        \hline
        d & 0, 1\\
        \hline
        p & 1, 2, 3, 4 \\
        \hline
        q & 1, 3, 5, 7, 9 \\
        \hline

    \end{tabular}
    \caption{\textbf{Arima Grid Search Parameters}: ranges of ARIMA hyperparameters to be searched during hyperparameter tuning.}
    \label{tab:ARIMA HPO}
\end{table}

\end{document}

\maketitle

\section{Results on 90-Day Forecast}

\begin{figure}[!htb]
\centering
\renewcommand{\figurename}{Supplementary Figure}
\includegraphics[width=0.5\linewidth]{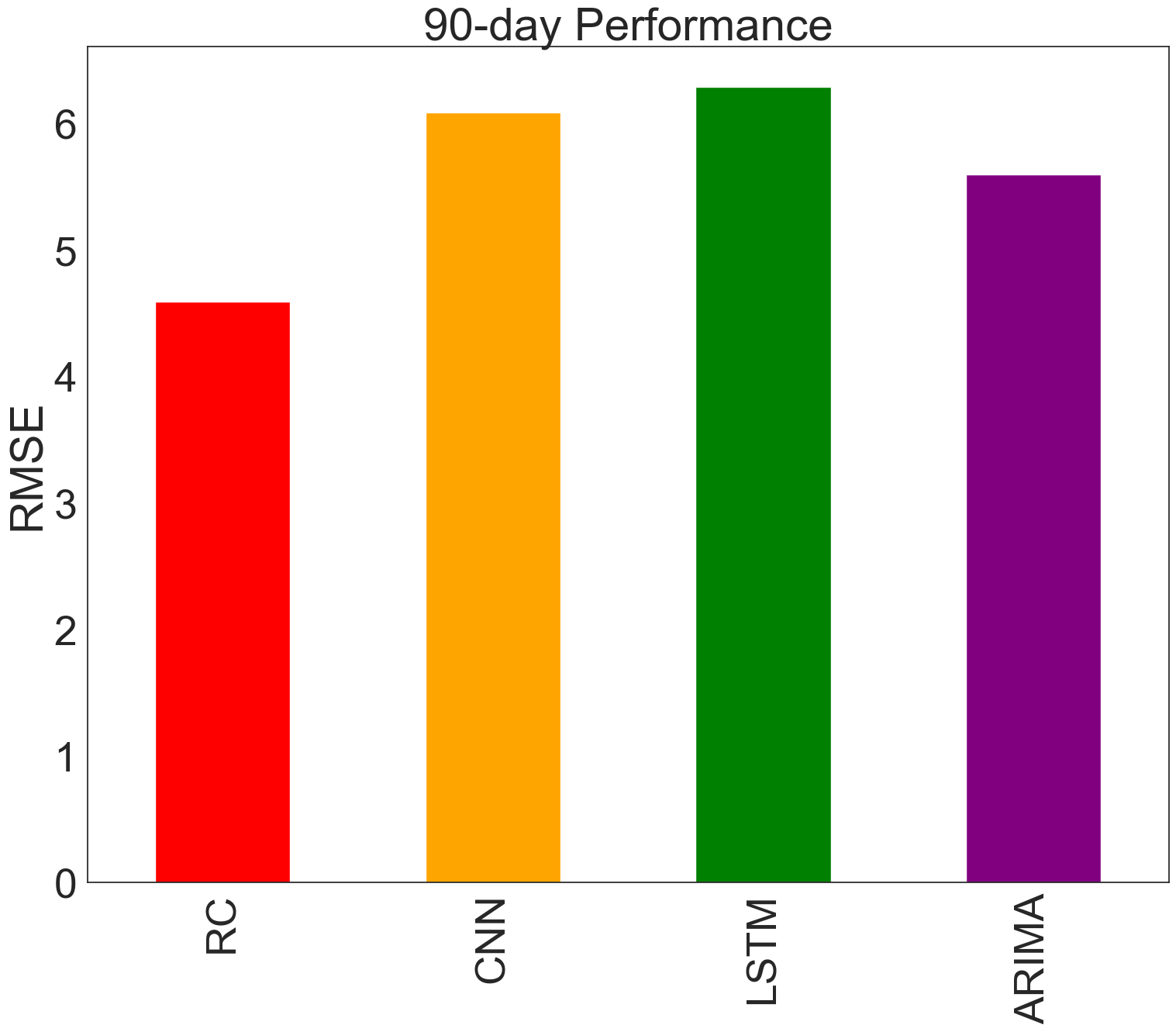}
\caption{Aggregated RMSE of the four methodologies on 90 day forecasting windows for the regions of Yemen\label{fig:90 days}}%
\end{figure}

In addition to our main analysis which was carried out on a 60-day window, we also tried forecasts on a 90 day window in the context of Yemen. As we can see in the aggregated results in figure \ref{fig:90 days} RC is confirmed as the best performing methodology even for this longer forecast horizon. Even though the numerical performances are good, we selected the 60-day window as our focus since the forecasts with all methodologies show a mean-reverting tendency for this window. This, despite not influencing the performances, limits the actual usability  of the forecasts, failing to signal deteriorating curves.

\section{Reservoir Computing Confidence Interval}

Confidence intervals for Reservoir Computing (RC) may be added by considering ±2-standard errors confidence bands (similarly to \cite{carammia2022forecasting}), where the standard error (SE) is computed as:

\begin{equation*}
    \text{SE} = \frac{\sqrt{\frac{1}{N} \sum_{i=1}^{N} ((y-y^i) - \mu)^2}}{\sqrt{N}}
\end{equation*}

\noindent
where $y-y^i$ is the residual error between the target value $y^i$ and the forecasted value $y$; $\mu$ is assumed to be 0, as we expect the population of residuals errors to be sampled from a normal distribution with 0 mean; and N is the number of out-of-sample splits for which these residuals are calculated.

\begin{figure}[!htb]
\centering
\includegraphics[width=1.1\textwidth]{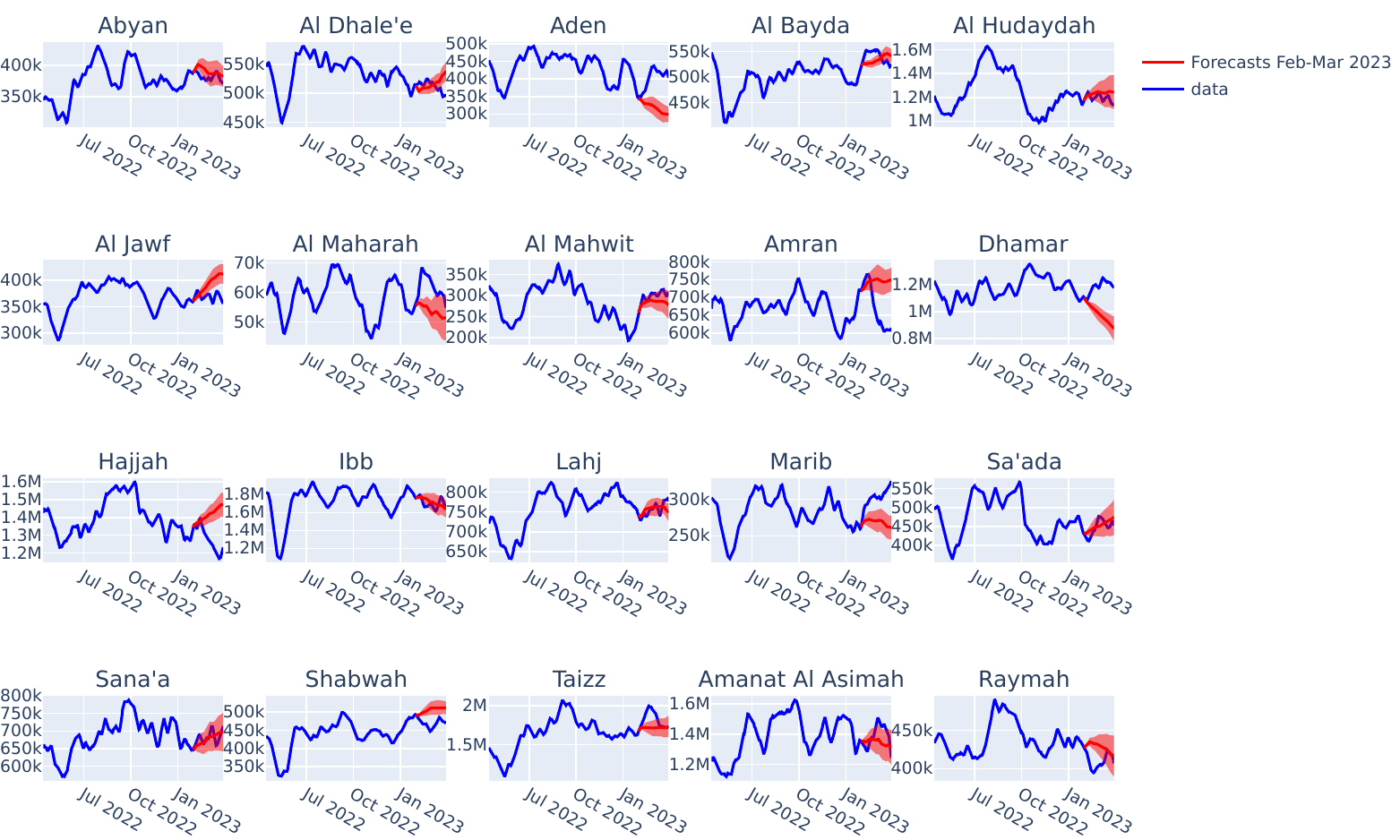}
\renewcommand{\figurename}{Supplementary Figure}
\caption{Example of confidence intervals computed using the standard error (SE) for forecasts in Yemen in February and March 2023
\label{fig:ci}}%
\end{figure}

Figure \ref{fig:ci} shows an example of forecasts with confidence intervals computed using ±2-standard errors confidence bands for each considered governorate in Yemen in February and March 2023. The residual errors are calculated for a testing set with 20 rolling splits in the 4 months preceding the window to be forecasted.

\section{Preliminary Results from Application}

\begin{figure}[!htb]
\centering
\includegraphics[width=\linewidth]{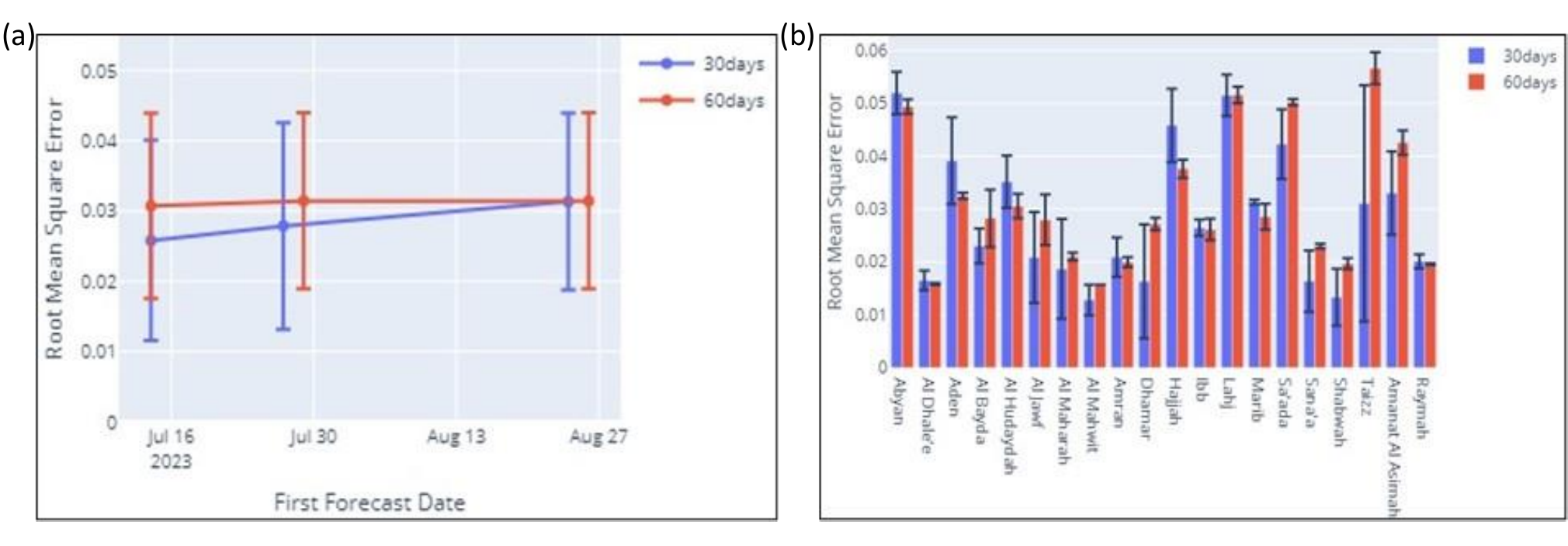}
\includegraphics[width=\linewidth]{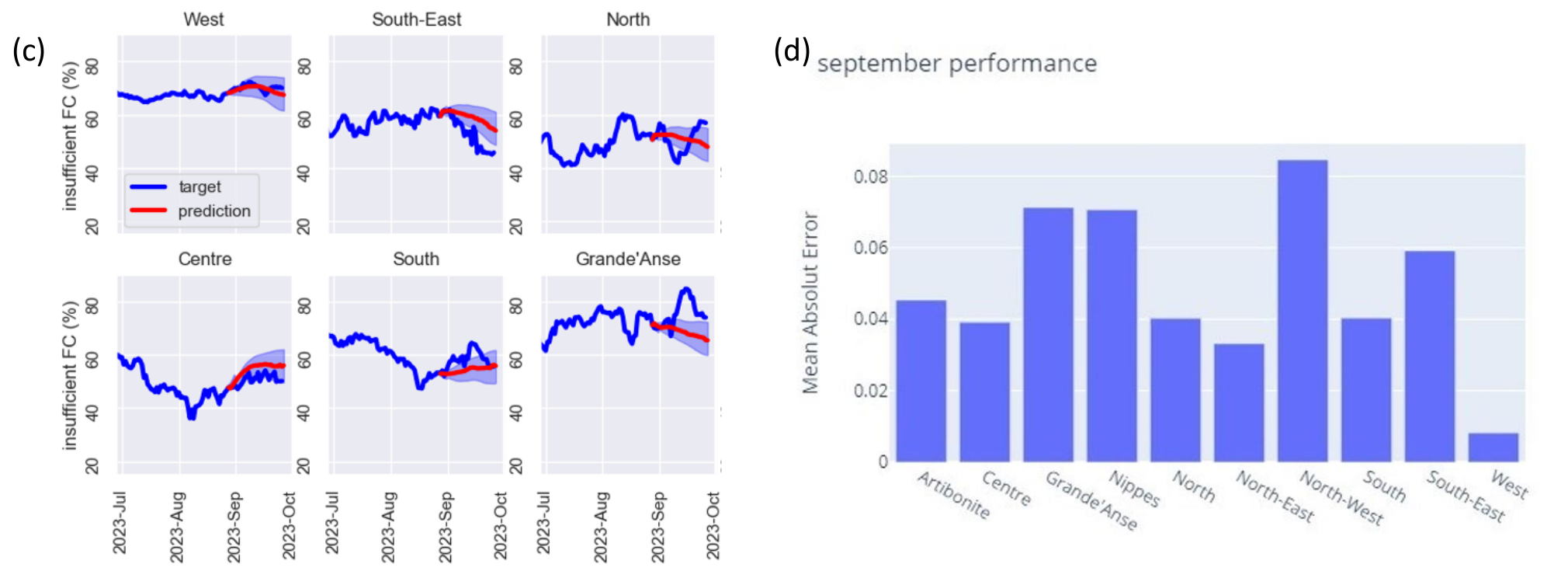}
\renewcommand{\figurename}{Supplementary Figure}
\caption{\label{fig:in production} The RMSE obtained when producing the forecasts for Yemen Country Office  \textbf{(a)-(b)} and Haiti \textbf{(d)}. These correspond the error of the Forecasting methodology in actual use and not on historical data. In \textbf{(c)} We show an example of the forecasted values (red curves) vs the actual data (blue curves). The confidence intervals represent the errors obtained on historical data.}%
\end{figure}

While the main text compares different methodologies on historical data where some aspects of a real-life application are not fully reproduced, the RC methodology developed in this work has already been applied as a prototype in collaboration with several WFP Country Offices. First evaluations of these forecasts can be found in figure \ref{fig:in production} for the context of Yemen and Haiti. In these cases delayed data sources have to be extrapolated with the last available value, shifted for training and prediction and, when provided, forecasts have been used. PEWI, NDVI and Rainfall come with associated forecasts. In extreme cases we drop features that are not available from the analysis.
While these analyses are carried out on a small sample they do not seem to indicate any significant performance drop which could have been caused by data delay. This is likely related to the importance of the auto-regressive component of the model we report in the manuscript. Another explanation is the importance of long term trends over short term fluctuations in time series like food- and headline inflation, which are also the most likely to experience a delay in reporting. 

\section{RC Hyperparameters}

\begin{figure}[!htb]
\centering
\includegraphics[width=\linewidth]{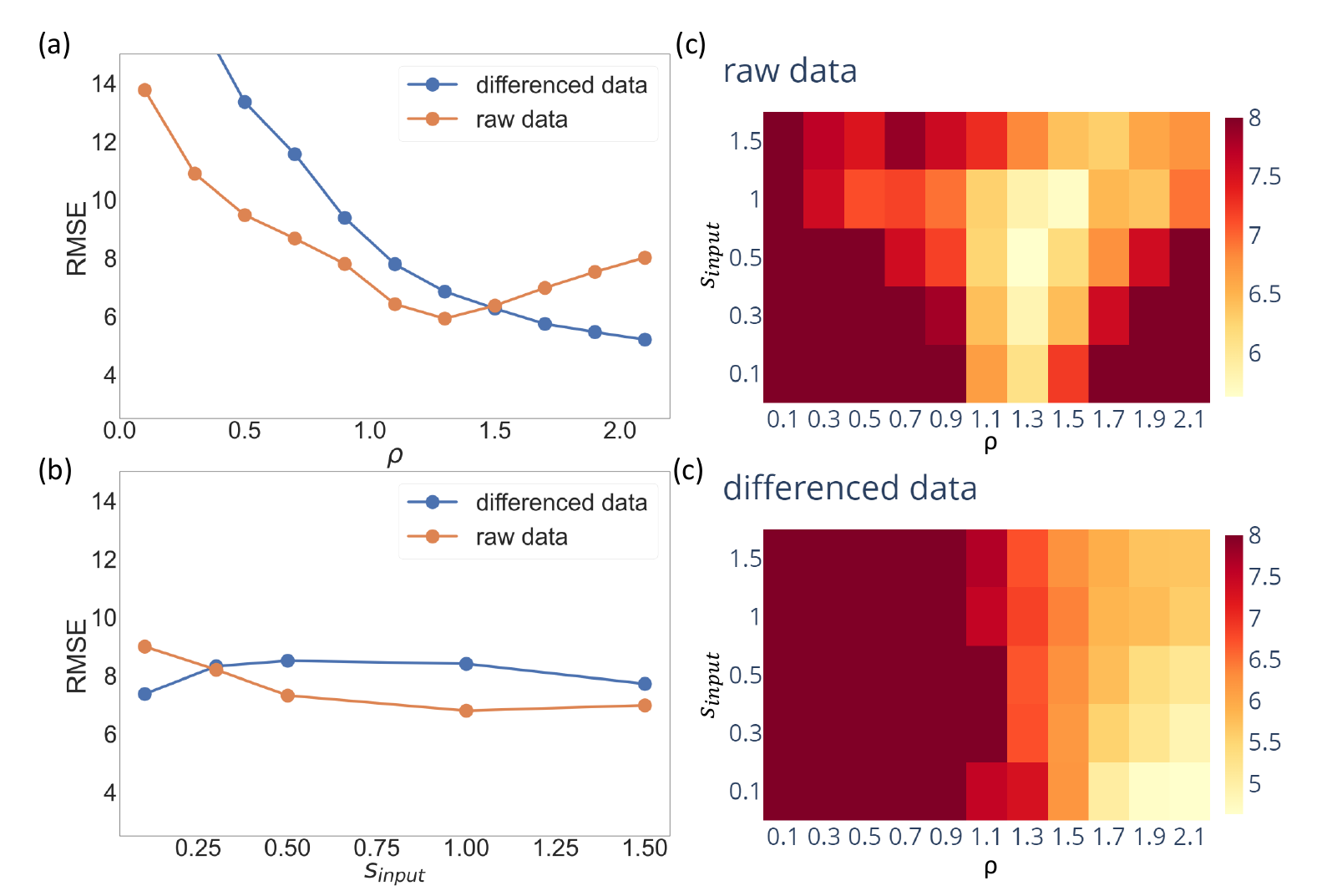}
\renewcommand{\figurename}{Supplementary Figure}
\caption{Influence of the hyperparameters spectral radius $\rho$, input strength $s_{input}$ and differencing on the RC model as extracted from the grid search results. The value is always the median RMSE over all splits and hyperparameters that are not specified. \textbf{(a):} The RMSE for different values of the spectral radius separated by application on differenced or raw data. \textbf{(b):} The RMSE for different values of the input strength separated by application on differenced or raw data. \textbf{(c):} Heatmap of the RMSE for different values of both hyperparameters applied on raw data. \textbf{(d):} Heatmap of he RMSE for different values of both hyperparameters applied on differenced data. \label{fig:grid}}
\end{figure}

The grid search we have carried out has given us a number of insights on the impact of hyperparameters beyond what has been reported in the manuscript. In particular we have seen that in the case of RC there are hyperparameter configurations that seem to lead to systematically good performances without needing adjustment for each split. In particular the hyperparameters of the spectral radius $\rho$, the input strength $s_{input}$ and the differencing lead to strong results and show very promising regimes. We show some of these results in figure \ref{fig:grid}.

As we can see the RC responds differently to the hyperparameters $\rho$ and $s_{input}$ depending on the use of differenced or raw data. While both approaches can be tuned to give good results, the differencing of the data seems to allow higher performances in particular in a regime of high $\rho$ and low $s_{input}$. This corresponds to a regime where the memory of the RC is particularly pronounced in contrast to recently incoming input. This indicates that the RC is making use of longer term trends instead of short term fluctuations. The hyperparameters indicated here can be used as a starting point for application if an in depth tuning is too expensive.

\section{Training Time and RC Ensemble Sizes}

\begin{figure}[!htb]
\centering
\includegraphics[width=\linewidth]{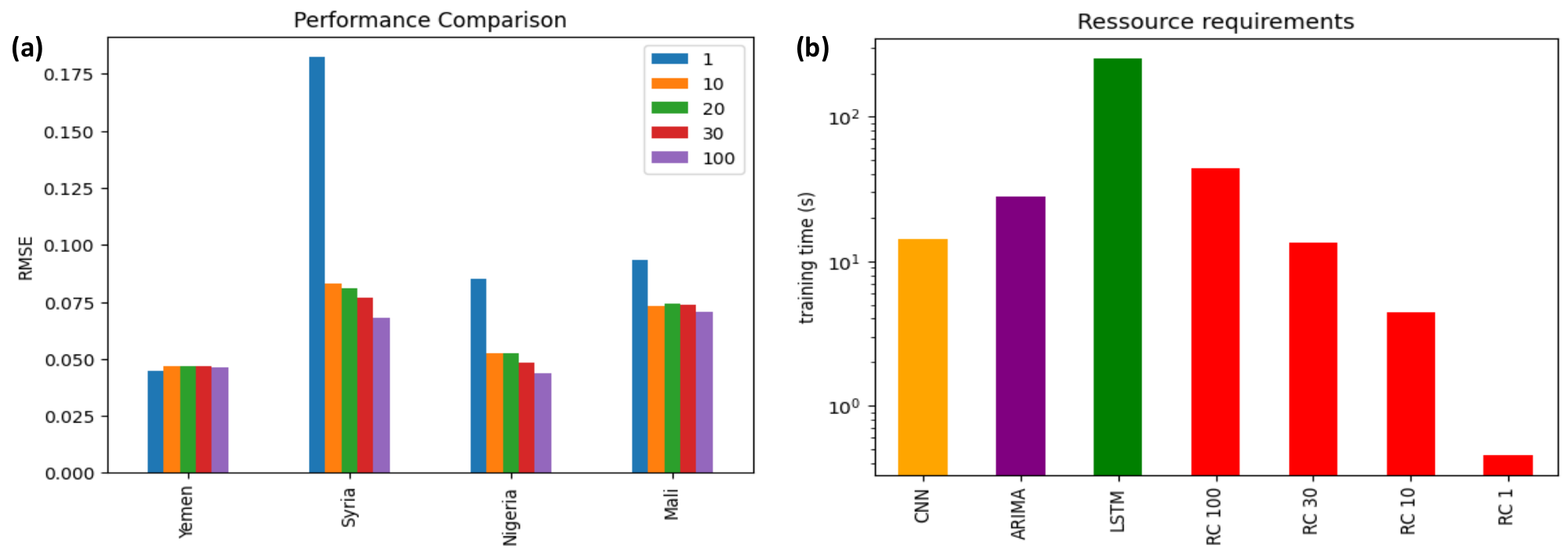}
\renewcommand{\figurename}{Supplementary Figure}
\caption{\textbf{(a)} Aggregate RMSE of Ensemble RC with 1, 10, 20, 30 and 100 models on the four analyzed countries. \textbf{b} The average training times required by the different algorithms including different ensemble sizes of RC.}
\label{fig:ensemble}
\end{figure}

Since RC is built on randomly selected parameters and is fast and efficient in its training we have used it as a building block for an ensemble model. This is meant to increase the forecasting power by stabilizing any fluctuations introduced through the influence of the stochasticity and making use of the extremely low training time of a single RC model. To validate our choice of using an ensemble of 100 we have tested the performance against ensembles of 1, 10, 20 and 30 on the the splits introduced in the main section using the hyperparameter configurations derived from the grid search. The results are shown figure \ref{fig:ensemble}(a).

We can see that a single RC model is often too unstable in this context, and  is especially evident in Syria where prediction mostly fails. However, even an ensemble of 10 models increases the performance drastically. A further increase up to 100 still leads to an overall improvement, but the performance seems to be slowly plateauing and while even larger ensembles might further increase the predictive power by a small amount, the added resource requirements are unlikely to be worth the improvement. A noteworthy outlier in this figure is the performance of a single RC model on Yemen, where it performs slightly better than the ensemble model and an increase of the ensemble size seems to add no value. This could be related to high amount and quality of the Yemen data. It is also important to note that the hyperparameters used in this comparison were originally tuned for an ensemble of 100, thus there is a bias present in this analysis. For example particularly unstable configurations might not have been chosen for a single model, but are still suitable for the ensemble.   

In figure \ref{fig:ensemble}(b) we observe the training time for all tested algorithms on a single split, using the hyperparameters selected from the grid search. For RC hyperparameters were optimized using 100 ensemble models. Excluding the RC with ensembles smaller than 100, the comparison reveals that the CNN achieves the shortest training time and also that, despite employing an ensemble of 100 models for the RC, its training time remains comparable to ARIMA and is significantly more efficient than LSTMs. The relatively long training times obtained using ARIMA depend on the necessity to train a distinct ARIMA model for each region in a country, in contrast to the other methodologies that enable simultaneous training allowing for concurrent forecasting across all regions. From this analysis it appears that all methodologies can be easily trained and used on a personal laptop and do not require a heavy infrastructure to be used operationally.

\section{Reservoir Computing and XGBoost}

\begin{figure}[!htb]
\centering
\includegraphics[width=1.1\textwidth]{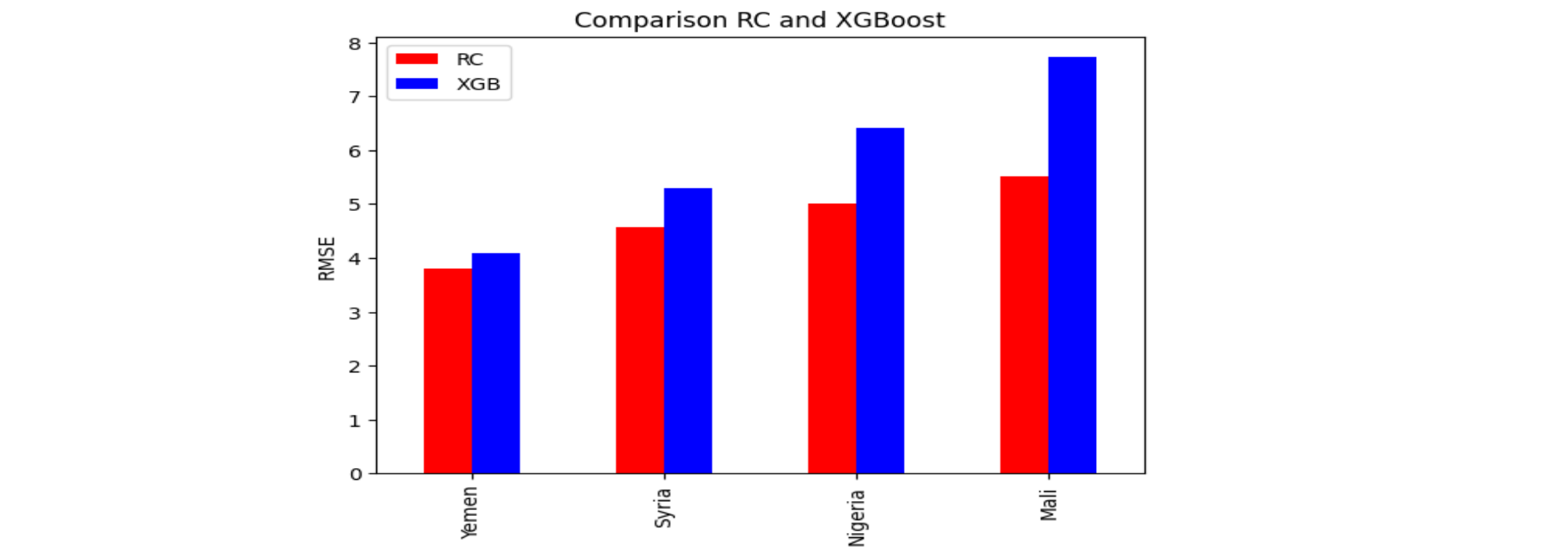}
\renewcommand{\figurename}{Supplementary Figure}
\caption{Comparison between the performances shown in \cite{foini2023forecastability} on 30-day forecasts and shared in the public repository of the publication. We can see that RC outperforms the XGBoost algorithm on all tested countries. }
\label{fig:Xgboost}
\end{figure}

To further evaluate our proposed approach, we conducted a benchmark on the same dataset employed and shared in \cite{foini2023forecastability}. In their study the authors implemented an XGBoost forecaster for a consecutive 30-day forecasting horizon, targeting the same variable as addressed in our manuscript. It's worth noting that XGBoost inherently lacks built-in support for multi-output regression, as its primary design focuses on single-output tasks such as regression and classification. Consequently, to achieve a 30-day prediction horizon, the authors were forced to train and evaluate thirty distinct models. In figure \ref{fig:Xgboost} we can clearly see how the RC outperformed the XGBoost in all tested countries. Given XGBoost's reliance on multiple models, rendering its use in a production environment challenging in terms of maintenance and updates, and the clear results obtained in comparing the performances on a horizon of 30 days, extending the approach to a $60$day prediction horizon was considered impractical and unlikely to provide substantial additional value.

\printbibliography